\documentclass[sn-nature]{sn-jnl}

\usepackage{graphicx}%
\usepackage{multirow}%
\usepackage{amsmath,amssymb,amsfonts}%
\usepackage{amsthm}%
\usepackage{mathrsfs}%
\usepackage[title]{appendix}%
\usepackage{xcolor}%
\usepackage{textcomp}%
\usepackage{manyfoot}%
\usepackage{booktabs}%
\usepackage{algorithm}%
\usepackage[frozencache,cachedir=minted-cache]{minted}

\usepackage{listings}%
\usepackage{booktabs}

\usepackage{ffcode}
\usepackage{adjustbox} 
\usepackage{fvextra}    
\usepackage{lmodern}

\usepackage{geometry}

\begin{document}

\title[Article Title]{Double Jeopardy and Climate Impact in the Use of Large Language Models: Socio-economic Disparities and Reduced Utility for Non-English Speakers}


\author*{\fnm{Aivin V.} \sur{Solatorio}}\email{asolatorio@worldbank.org}
\equalcont{These authors contributed equally to this work.}

\author{\fnm{Gabriel} \sur{Stefanini Vicente}}
\equalcont{These authors contributed equally to this work.}

\author{\fnm{Holly} \sur{Krambeck}}

\author{\fnm{Olivier} \sur{Dupriez}}

\affil{\orgname{The World Bank}, \orgaddress{\street{1818 H Street N.W.}, \city{Washington}, \postcode{20433}, \state{District of Columbia}, \country{USA}}}


\abstract{Artificial Intelligence (AI), particularly large language models (LLMs), holds the potential to bridge language and information gaps, which can benefit the economies of developing nations. However, our analysis of FLORES-200, FLORES+, Ethnologue, and World Development Indicators data reveals that these benefits largely favor English speakers. Speakers of languages in low-income and lower-middle-income countries face higher costs when using OpenAI's GPT models via APIs because of how the system processes the input---tokenization. Around 1.5 billion people, speaking languages primarily from lower-middle-income countries, could incur costs that are 4 to 6 times higher than those faced by English speakers. Disparities in LLM performance are significant, and tokenization in models priced per token amplifies inequalities in access, cost, and utility. Moreover, using the quality of translation tasks as a proxy measure, we show that LLMs perform poorly in low-resource languages, presenting a ``double jeopardy" of higher costs and poor performance for these users. We also discuss the direct impact of fragmentation in tokenizing low-resource languages on climate. This underscores the need for fairer algorithm development to benefit all linguistic groups.}


\maketitle

\section{Introduction}\label{introduction}

Given their transformative impact on society, it is imperative to investigate the potential inequalities stemming from large language models. Like previous general-purpose technologies such as the steam engine, electricity, and the Internet, LLMs can significantly alter economies, cultures, and social frameworks \cite{crafts_artificial_2021}. However, they also have the capacity to intensify existing disparities or generate new ones. LLMs are trained on extensive corpora of Internet data that mirror the global digital footprint \cite{penedo_refinedweb_2024}. While these models compress vast textual information to form representations of the world, their dependence on such data subjects them to inherent biases, particularly those influenced by the digital divide \cite{khowaja_chatgpt_2024}. As a result, discrepancies in Internet content coverage lead to LLMs having an uneven impact on different demographic groups.

Ensuring equitable access to technology is essential. Yet, specific technical barriers, like the tokenizer---an integral part of LLMs responsible for processing input texts---impede this objective. Tokenizers, much like LLMs, are trained on text corpora to learn the distribution of syntactic fragments or tokens. Customizing tokenizer training to particular corpora improves its application effectiveness by pre-learning syntactic characteristics and enhancing processing efficiency \cite{dagan_getting_2024,ali_tokenizer_2024}. However, prior research has demonstrated that tokenizers employed by services such as OpenAI create inequalities for non-English speakers \cite{petrov_language_2024}.

When processed by tokenizers, non-English languages often break down into more tokens than English---this is referred to as fragmentation \cite{petrov_language_2024}. This issue underscores the lack of linguistic diversity in the datasets used to train these tokenizers. Since a single token is typically the unit of pricing for most paid LLM API services, languages that experience greater fragmentation will incur higher costs \cite{dong_large_2024}. Thus, addressing linguistic diversity in algorithmic development is crucial to ensure equitable access to technology.

The socioeconomic effects of LLMs are still not well quantified. However, existing studies suggest that languages spoken primarily in countries with a lower Human Development Index (HDI) face more significant fragmentation \cite{ahia_all_2023}. This indicates that LLMs may impart uneven impacts influenced by geographical and socioeconomic factors since elevated fragmentation directly translates to increased usage cost.

The climate impact of LLMs has also been investigated \cite{bender_dangers_2021,faiz_llmcarbon_2023,rojas_evaluating_2024,li_toward_2024}. As demand for more powerful AI systems increases, the corresponding emissions associated with training and the use of models are concerning \cite{strubell_energy_2020,kerr_ai_2024}. Although AI and LLM are promising, their impact on climate remains grim \cite{luccioni_power_2024}. In addition, the discussion of how AI can impact water resources has also been studied \cite{li_making_2023}. Fortunately, large companies that conduct model training have net zero commitments and are taking steps to offset carbon emissions from these activities \cite{smith_microsoft_2020,meta_climate_nodate}.

\begin{figure}[tbp]
    \begin{center}
        \centerline{\includegraphics[width=0.97\columnwidth]{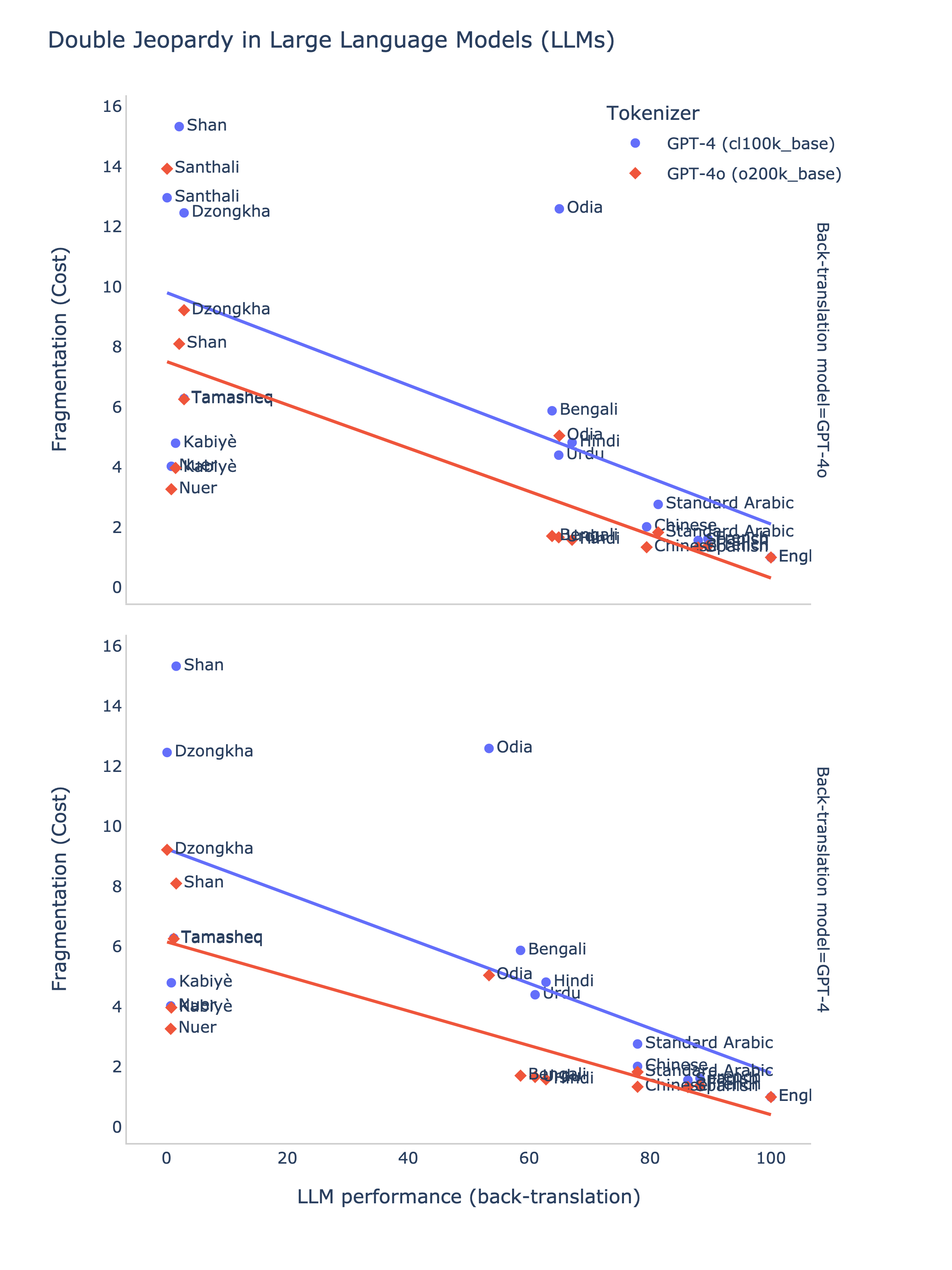}}
        \caption{The figure summarizes the double jeopardy in low-resource languages---such as Shan, Santhali, Dzongkha, Tamasheq, Kabiyè, Nuer---mostly spoken in low- and lower-middle-income countries. The cost of using LLMs is higher for these languages when the pricing is based on tokenization. The performance of LLMs in these languages is also poor. This shows results using tokenizers for GPT-4 and GPT-4o. The trendlines suggest that the GPT-4o tokenizer has generally reduced fragmentation. Derivation of the values used in this figure is detailed in Section~\ref{method}.}
        \label{fig:double-jeopardy-llm}
    \end{center}
\end{figure}

This research adds to the growing body of evidence on the socioeconomic challenges posed by LLMs, emphasizing that languages spoken by economically disadvantaged populations face a ``double jeopardy": higher costs for LLM usage and lower performance outcomes. Speakers of these languages are disproportionately affected by tokenization inefficiencies, which inflate the usage cost while delivering suboptimal results. A summary of this phenomenon for select languages is illustrated in Figure~\ref{fig:double-jeopardy-llm}. Furthermore, drawing on the existing literature, we connect the issue of fragmentation with its environmental consequences, demonstrating that the increased computational demands associated with the processing of fragmented languages lead to higher carbon emissions. Also, to our knowledge, this is the first work to quantify the population affected by these disparities in LLM performance and cost.

Finally, we examine the impact of unlocalized access costs for LLMs, which creates an inverse geographic arbitrage that disproportionately disadvantages certain populations. We also address the potential for further degradation in LLM performance as Internet data become increasingly polluted by LLM-generated content.

\section{Data}\label{data}

This section offers a detailed overview of the different data sources utilized in our analysis.

\subsection{FLORES-200 and FLORES+}

We use the FLORES dataset, an extensive collection of concise excerpts drawn from Wikipedia articles on various topics. Each excerpt has been translated into all included languages, making this dataset particularly valuable for multilingual research and applications such as machine translation, natural language processing, and linguistic studies. We utilize a combination of the FLORES-200 dataset \cite{team_no_2022} and the FLORES+ dataset \cite{noauthor_oldi_nodate}, both curated by the Open Language Data Initiative. The FLORES+ dataset includes 12 additional language variants compared to the original FLORES-200. Hereafter, we will refer to the combined dataset as FLORES-200P.

\subsection{Ethnologue}

We supplemented the FLORES-200P with data from the Ethnologue platform \cite{lewis_ethnologue_nodate}. Ethnologue provides extensive information on global languages. We gathered data on the estimated number of speakers, the level of digital language support, and the language family. Additionally, we collected information on the countries where each language is spoken and the respective number of speakers. This enabled us to study the relationship between countries and their spoken languages. Overall, we have successfully gathered data for 194 out of the 200 languages in the FLORES-200P dataset.

\subsection{World Bank World Development Indicators}

We use data from the World Bank’s World Development Indicators (WDI) to add socioeconomic insights \cite{noauthor_world_nodate}. For comparing economic performance and living standards across countries, we use the GDP per capita in current US\$ indicator (\texttt{NY.GDP.PCAP.CD}). The annual population growth rate (\texttt{SP.POP.GROW}) helps standardize the number of speakers per language from Ethnologue. Additionally, we obtained the World Bank’s country classification by income level through their Indicators API \cite{noauthor_world_nodate-1}.

\section{Anatomy of Large Language Models}\label{anatomy-llm}

Prior to our analysis, this section provides a brief synopsis of the key elements of LLMs---specifically, the transformer architecture. Additionally, we will explore the tokenization process and its role in perpetuating inequality within LLMs, especially affecting low-resource languages. This inequity poses notable challenges for non-English speakers, which will be thoroughly examined in this paper.

\subsection{Transformer Architecture}

The core of LLMs is the transformer architecture, which has revolutionized the processing and modeling of sequential data \cite{vaswani_attention_2017}. Unlike traditional recurrent neural network (RNN) models, transformers use a self-attention mechanism to assess relationships between all tokens in a sequence at once, rather than in order. This advancement helps in capturing long-range dependencies better, greatly improving performance across various natural language processing (NLP) tasks.

LLMs, typically structured as decoder-only transformers like Generative Pre-trained Transformers (GPTs), are trained to predict the next token in a sequence \cite{radford_improving_2018}. These models consist of multiple transformer blocks with multi-head self-attention and position-wise feedforward networks. Self-attention assigns importance scores to each token, helping the model grasp context and meaning. Residual connections and layer normalization within each block ensure stable training. This design enhances computational efficiency through parallelization and achieves high accuracy and coherence in language modeling tasks.

By training on extensive text corpora, these models discern underlying language patterns and relationships, which enables them to generate coherent, contextually appropriate text. The transformer architecture has revolutionized natural language processing, fostering the creation of advanced language models that emulate human abilities in text comprehension and generation. Its proficiency in managing long-range dependencies and its suitability for parallel processing have greatly enhanced model performance and scalability, propelling progress in applications like conversational agents, automated summarization, and language translation.

\subsubsection{Input Embeddings}

A transformer model's input is usually a sequence of tokens from the tokenization process, with each token represented as a vector. Positional embedding is added to include token positions in the sequence, forming an effective initial input for the model. These token embeddings, learned during training, encapsulate the semantic and syntactic details of each token, aiding in text comprehension and processing. The embeddings then go through transformer blocks for transformations that capture their complex interrelationships.

\subsubsection{Self-attention Mechanism}

The self-attention mechanism allows the model to assess the importance of each token in a sequence relative to the others. It computes a weighted sum of the token representations, where the weights represent how relevant each token is to the others in the sequence. This enables the model to dynamically focus on different parts of the input, capturing long-range dependencies between tokens and improving its understanding of context.

\subsubsection{Feedforward Neural Network Layers}

The feedforward neural network layers refine the output from the self-attention mechanism by applying nonlinear transformations, enabling the model to capture more complex patterns in the data. Each transformer block includes residual connections and layer normalization, which enhance training stability and improve overall model performance by preserving important features and avoiding vanishing gradients.

\subsection{Tokenization}

Tokenization is crucial for transformer models to handle text data properly. It breaks down text into tokens, which are then turned into numerical representations. This process is vital for the model to interpret and produce text precisely, affecting the detail it can capture.

Tokenizers, specialized algorithms, split raw text into tokens that can be words, subwords, or characters \cite{devlin_bert_2019}. Subword tokenization, commonly used in LLMs, strikes a balance between vocabulary size and the representation of rare words, enhancing the model's ability to handle out-of-vocabulary words and morphological variations. An example of this is Byte-Pair Encoding (BPE) \cite{brown_language_2020}. Customized tokenizers for other LLM applications have also shown improved performance \cite{solatorio_realtabformer_2023,solatorio_geoformer_2023}.

The tokenizer splits the text based on predefined rules or learned patterns, creating structured input for the model. The text may undergo preprocessing like lowercasing, punctuation removal, and handling special characters to normalize it, though sometimes it is processed as-is. Each token receives a unique identifier mapped to an embedding vector, which acts as a dense numerical representation carrying its semantic information for the model's processing.

\begin{figure}[tbp]
    \centering
    \includegraphics[width=0.97\textwidth]{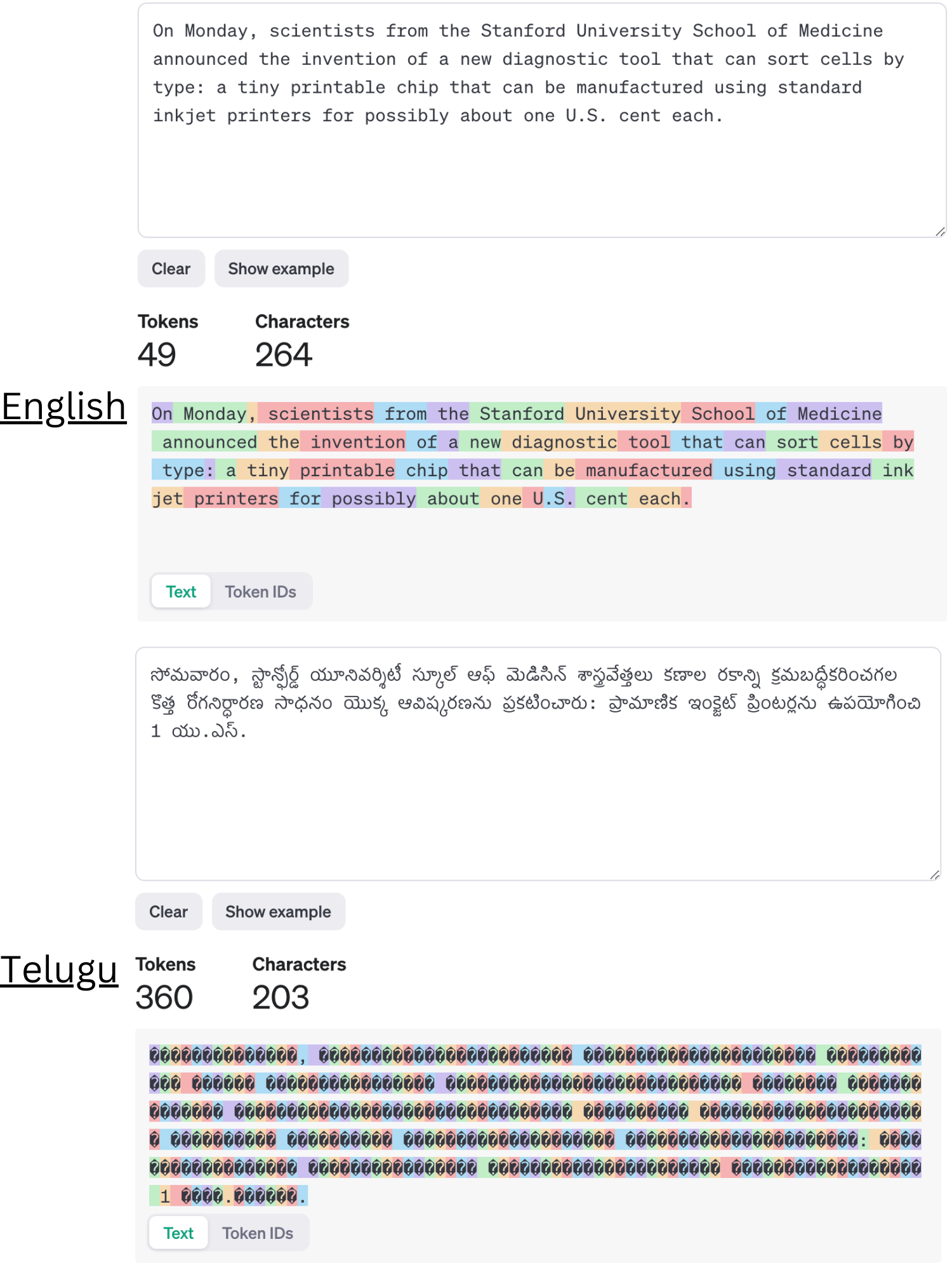}
    \caption{Visualization of tokenization for an equivalent sentence in English and Telugu (\url{https://platform.openai.com/tokenizer}). Note the number of tokens for each language after applying the tokenizer. Despite Telugu having fewer characters than the English equivalent, English only has 49 tokens, while Telugu resulted in 360 tokens. A fragmentation rate of around seven times.}
    \label{fig:eng-tel-gpt-4-tokenizer}
\end{figure}

Tokenization can be particularly challenging for languages with complex structures or limited resources, as shown in Fig.~\ref{fig:eng-tel-gpt-4-tokenizer}. Ineffective tokenization can harm model performance, so it is vital to optimize this process for better accuracy and efficiency in transformer models.

\section{Methodology}\label{method}

This section details our methods for creating linguistic and socio-economic indicators to measure economic disparities in access to enterprise LLM technologies between non-English and English languages. We propose a metric for assessing tokenization fragmentation costs across languages and outline our approach for evaluating LLM performance in different languages. Combining these indicators, we highlight the challenges non-English speakers face in accessing and using LLMs.

\subsection{Linguistic and Socio-economic Indicators}

To measure the economic disparity in access to language technologies, we employ a range of linguistic and socio-economic indicators. First, we outline our strategy for normalizing the number of speakers per language. Subsequently, we detail our approach to developing a weighted wealth of linguistic indicators and associating languages with income levels, thereby establishing a connection between language and economic factors.

\subsubsection{Number of Speakers for Each Language}

Ethnologue is an extensive database offering in-depth details about languages spoken globally \cite{lewis_ethnologue_nodate}. It notably includes data on the number of speakers per country for each language, although this information might come from various sources and different periods.

\begin{figure}[t]
\begin{center}
\centerline{\includegraphics[width=\columnwidth]{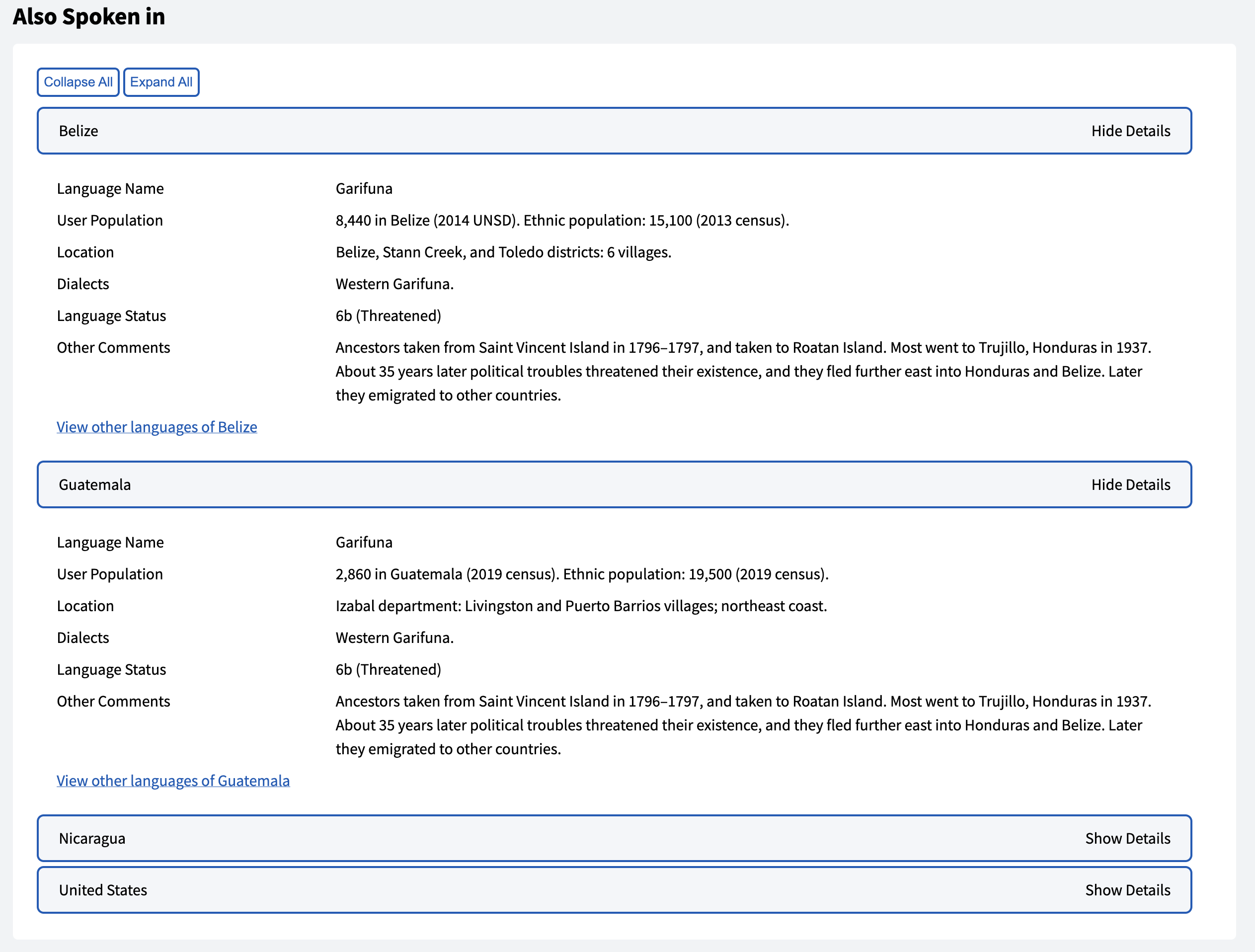}}
\caption{Ethnologue page showing the various locations where the Garifuna language—mainly spoken in Honduras—is also spoken. The population data are reported based on different sources and points in time.}
\label{fig:eth-lang-page}
\end{center}
\end{figure}

The information accessible on the platform for a specific language is illustrated using the Garifuna language as an example, shown in Fig.~\ref{fig:eth-lang-page}. Although the number of speakers in countries where Garifuna is spoken is recorded, the data sources lack consistency. For instance, the figures for Belize are based on 2014 data from the United Nations Statistics Division (UNSD), whereas the number of Garifuna speakers in Guatemala is derived from the 2019 census. To address this inconsistency, we suggest a method to standardize the reported number of speakers by factoring in population growth rates.

To compare language speakers in FLORES-200P accurately, we adjust Ethnologue's reported counts to a recent timeframe using country-specific population growth rates. For speaker numbers from 2022 or later, we keep the original value. If the data is before 2022, we project it to 2022 using the World Bank's latest annual population growth rate indicator (\texttt{SP.POP.GROW}). If no date is available, we leave the count as is. The adjustment process is formalized below:

\begin{equation}\label{eq:speaker-adjustment}
    S_{i,j} = S_{i,j,t} \times
    \begin{cases}
        1                                  & \text{if } t \geq 2022 \text{ or } t \text{ is unavailable} \\
        \prod_{k=t+1}^{2022} (1 + g_{j,k}) & \text{if } t < 2022
    \end{cases}
\end{equation}

where, \(S_{i,j}\) denotes the adjusted number of speakers of language \(i\) from country \(j\), while \(S_{i,j,t}\) represents the number of speakers of language \(i\) from country \(j\) in year \(t\) as reported by Ethnologue. The variable \(g_{j,k}\) stands for the annual population growth rate for the country \(j\) reported at year \(k\).

The overall number of speakers for each language is determined by adding up the adjusted speaker counts from all the countries where the language is used.

\begin{equation}\label{eq:total-speakers}
    S_{i} = \sum_{j=1}^{n} S_{i,j}
\end{equation}

This method has challenges, mainly assuming that language speaker growth in each country mirrors overall population growth. This isn't always true; a language's speakers may decline even if the population rises. We also didn't differentiate between primary and secondary language speakers due to data limitations, leading to potential overcounts when summing languages. Despite these issues, we believe this provides a reasonable approximation of 2022 language speakers, improving cross-language comparisons. These limitations highlight the need for more detailed data on language speakers to improve accuracy.

\subsubsection{Estimate GDP per Capita for each Language}

To incorporate an economic aspect into our study, we establish a measure to assess the wealth tied to each language. Using the harmonized speaker counts, $S_{i,j}$, calculated earlier in Eq.~\ref{eq:speaker-adjustment}, we consider languages spoken in various countries.

To calculate a population-weighted GDP for each language, we use the per capita GDP in current USD (\texttt{NY.GDP.PCAP.CD}) for each country. This approach offers a detailed measure of the economic impact tied to each language, enabling accurate comparisons of economic contributions among languages. The indicator is calculated as follows:

\begin{equation}\label{eq:wealth-indicator}
    W_i = \frac{\sum_{j=1}^{n} S_{i,j} \times \text{GDP}_{j}}{\sum_{j=1}^{n} S_{i,j}}
\end{equation}

where \(W_i\) is the population-weighted GDP for language \(i\), \(S_{i,j}\) is the adjusted number of speakers of language \(i\) from country \(j\) as per Eq.~\ref{eq:speaker-adjustment}, and \(\text{GDP}_{j}\) is the per capita GDP in USD for country \(j\) according to the World Bank’s World Development Indicators API for 2022. To calculate the population-weighted GDP for each language, we sum the products of adjusted speaker counts and per capita GDP across all relevant countries, then normalize by dividing by the total number of adjusted speakers.

\subsubsection{Classify Income Level for Each Language}

Languages traverse boundaries, reaching from individual speakers to entire regions and countries. The World Bank Group categorizes the world's economies into four income levels: low, lower middle, upper middle, and high \cite{noauthor_world_nodate-2}. To proportionately reflect the income level of language speakers, we use this classification to define an indicator that represents the population-weighted average income level of the countries where the language is spoken, on a scale from 0 (low) to 1 (high). For example, a value of 1 would mean that all countries where a particular language is spoken are classified as high-income. The income level classification for each language is derived as follows.

\begin{equation}\label{eq:income-level}
    \text{I}_{i,j} = \frac{\sum_{k} S_{i,j,k}}{\sum_{m} \sum_{k} S_{i,m,k}}
\end{equation}

where \(\text{I}_{i,j}\) denotes the population-weighted income level factor for language \(i\) and income level \(j\), and \(S_{i,j,k}\) represents the adjusted number of speakers of language \(i\) in country \(k\) with income level \(j\). The numerator computes the total number of speakers for language \(i\) in countries with income level \(j\), whereas the denominator normalizes this by summing speakers across all income levels and countries.

Another wealth-based language classification can be derived from Eq.~\ref{eq:wealth-indicator} using the thresholds used to calculate the income classification. Note that the income thresholds are intended to be applied to GNI values, but the value presented in Eq.~\ref{eq:wealth-indicator} uses the GDP. Since GNI is calculated by adding net income from abroad to the GDP, using the GDP-based values will tend to lower the classification. However, we choose the GDP because we do not want to account for external wealth. The thresholds used are low income ($<$1,145), lower-middle income (1,146 - 4,515), upper-middle income (4,516 - 14,005), and high income ($>$14,006).

\subsection{Quantifying Fragmentation Cost}

Lastly, we describe the framework for assessing fragmentation cost in LLM systems and detail our approach for evaluating LLM performance across different languages, particularly in translation tasks.

\subsubsection{Tokenization Premium per Language}

We use the concept of \textit{premium} from \cite{petrov_language_2024} to systematically assess how tokenizers process equivalent sentences in various languages. Consider sentence \( s_A \) in language A and its translation \( s_B \) in language B. The ratio

\begin{equation}
    P_{A,B} = \frac{\|t(s_A)\|}{\|t(s_B)\|}
\end{equation}

measures the \textit{premium} of A compared to B, where \( t(\cdot) \) signifies a tokenizer and \( t(s_A) \) indicates the tokenization of sentence \( s_A \), with \( \|t(s_A)\| \) representing its length.

Although calculating the premium is feasible for any language pair, we concentrate on comparing each language to English. English being the most prevalent language in LLMs and a standard for tokenization informs this choice. Thus, the premium for language \( l \) relative to English is simplified as follows:

\begin{equation}\label{eq:fragmentation}
    P_{l} = \frac{\|t(s_l)\|}{\|t(s_{\text{English}})\|}
\end{equation}

This indicator highlights differences in tokenization efficiency across languages and related costs compared to English. A higher premium means less efficient processing, leading to more fragmentation and potentially higher user costs, worsening socioeconomic disparities. By measuring this premium, we can identify which languages struggle with tokenization and evaluate its impact on LLM accessibility and performance, guiding efforts to improve tokenization strategies.

\subsection{FLOPs as an Indicator of the Climate Impact of LLMs}\label{subsec:flops}

FLOP count, or floating point operations, is one of several components used to estimate the carbon emissions associated with LLMs. Other key factors include the energy efficiency of the hardware, the cooling and infrastructure of the data centers, and the energy mix, whether the electricity that powers the computations comes from renewable or non-renewable sources \cite{li_toward_2024,faiz_llmcarbon_2023}. Among these, the FLOP count serves as a valuable proxy indicator for the computational requirements of LLMs, offering insights into their potential climate impact.

An indicator for estimating the inference cost of LLMs in terms of FLOP count has been presented in the seminal paper on the scaling laws of LLMs \cite{kaplan_scaling_2020} as well as in the recent literature attempting to quantify a holistic computation of carbon emissions related to the technology \cite{faiz_llmcarbon_2023}. It is approximated as follows:

\begin{equation}\label{eq:flop}
    F_{IC} \approx 2PD
\end{equation}

where $F_{IC}$ is the estimated FLOP count, $P$ represents the count of non-embedding parameters in the LLM, while $D$ is the number of tokens processed. The dense parameter count is used for LLMs implemented as a mixture of agents (MoE). In the accounting model presented in \cite{faiz_llmcarbon_2023}, the FLOP count is directly proportional to the total carbon emission. This accounting allows us to mathematically show that fragmentation directly contributes to the climate impact of LLMs since $D$ is the total number of tokens processed.

\subsection{Translation Tasks as a Proxy for LLM Performance}\label{sec:translation-proxy-llm}

The indicators discussed earlier aim to reveal the inequalities in LLMs due to tokenizer choices, leading to fragmentation. As LLM usage is priced per token, these differences cause disparities. This section examines another inequity: the model's ability to perform tasks effectively across various languages.

We use language translation to evaluate LLMs' performance across different languages, focusing on low-resource languages. Although past assessments have been done \cite{peng_towards_2023,zhu_multilingual_2024}, they rarely target our languages of interest. We designed an experiment to test the LLM's ability to translate low-resource languages into English, serving as a measure of its multilingual capabilities.

\subsubsection{Translation Task}

We select various languages from the FLORES-200P dataset to evaluate the translation performance of the LLM. Our selection focuses on languages primarily spoken in low- and lower-middle-income countries, as well as those widely spoken globally. This diverse array covers multiple linguistic families and regions, ensuring a thorough assessment of the LLM’s translation capabilities.

We take sentences from the FLORES-200P dataset's selected source languages and have the LLM translate them into English. Translating at the sentence level maintains the independence of each translation. To keep the experiment consistent, we use a basic system prompt. While improved performance through prompt engineering is documented \cite{yamada_optimizing_2023,he_prompting_2024}, we choose simplicity to avoid any prompt influence on the LLM's performance. The system prompt is detailed in Listing~\ref{lst:system-prompt}.

\begin{listing}[ht]
    \begin{minted}[breaklines, breaksymbolleft={}]{text}
    You are a highly advanced machine translation system specializing in translations from {source_language} to English. Please translate the given text by the user, and format your response as follows: `English: <translation>`.

    Provide a high-quality translation that accurately conveys the meaning of the original text.
    \end{minted}
    \caption{System prompt used for the translation task from the source language to English.}
    \label{lst:system-prompt}
\end{listing}

After giving the system prompt, we input the individual sentences directly into the LLM without further instructions. Furthermore, we incorporate insights from previous research to configure LLM parameters in the translation task \cite{peng_towards_2023}. In particular, we use a temperature setting of 0, which has been determined to yield the best performance. We also identify the prefix `English: ' to parse the translation output from the LLM.

\subsubsection{Measuring Translation Quality}

Once the sentences are translated, we have the LLM compare them to their original English counterparts. We first evaluate translation quality through a binary classification task, where the LLM identifies each translation as correct or incorrect. This assessment helps measure the LLM's effectiveness in translating from the source language to English.

We developed two different versions of this evaluation approach to ensure robustness. The initial version simply prompts the LLM to decide whether the translation is accurate or not---a form of zero-shot prompting that relies heavily on the intrinsic reasoning capabilities of the LLM.

The second method includes a chain-of-thought component, where the LLM first explains its decision before delivering a verdict. This approach has been found to improve the quality of reasoning in LLM \cite{wei_chain--thought_2024}. The zero-shot prompt is shown in Listing~\ref{lst:system-prompt-binary-no-explanation} and the chain-of-thought prompt in Listing~\ref{lst:system-prompt-binary-explanation}.

\begin{listing}[ht]
    \begin{minted}[breaklines, breaksymbolleft={}]{text}
    You are an expert machine translation evaluation system, capable of accurately assessing precise matches between original and translated texts. Given an original English sentence and its back-translation into English from another language, assess whether the retranslated sentence accurately conveys the same meaning as the original, ensuring that all facts and details are preserved.

    Rate the translation quality as either `CORRECT` if the translated sentence is semantically identical to the original, preserving all factual information and details, or `INCORRECT` if it differs in meaning, omits or distorts any facts or details.

    Respond with: `Rating: <rating>`. Provide no further explanation.
    \end{minted}
    \caption{System prompt used to assess the correctness of the translated sentence using a binary classification. The LLM is instructed to directly provide a rating without further explanation.}
    \label{lst:system-prompt-binary-no-explanation}
\end{listing}

\begin{listing}[ht]
    \begin{minted}[breaklines, breaksymbolleft={}]{text}
    You are an expert machine translation evaluation system, capable of accurately assessing precise matches between original and translated texts. Given an original English sentence and its back-translation into English from another language, assess whether the retranslated sentence accurately conveys the same meaning as the original, ensuring that all facts and details are preserved.

    Rate the translation quality as either `CORRECT` if the translated sentence is semantically identical to the original, preserving all factual information and details, or `INCORRECT` if it differs in meaning, omits or distorts any facts or details.

    First, explain to yourself in one sentence the reason for your rating. Then, end your response with `Rating: <rating>`.
    \end{minted}
    \caption{System prompt used to assess the correctness of the translated sentence using a binary classification. The LLM is instructed to first provide an explanation before providing the rating.}
    \label{lst:system-prompt-binary-explanation}
\end{listing}

Besides the two binary prompting tasks, we use a five-point rating scale—Poor, Fair, Good, Very Good, and Excellent---for a more detailed evaluation. The LLM rates translations on this scale, which helps in further assessing its translation performance.

Here, we rely on chain-of-thought prompting since it tends to perform better according to research. Also, the prompt includes descriptions for each rating scale, giving context and grounding to the LLM's assessment. The descriptions offer further guidance for the qualitative review of the LLM's verdict.

\begin{listing}[H]
    \begin{minted}[breaklines, breaksymbolleft={}]{text}
    You are an expert machine translation evaluation system, capable of accurately assessing translation quality. Given a source text and its translated counterpart, rate the translation quality using a 5-point scale: Poor, Fair, Good, Very Good, Excellent. The scale is defined as follows:

    **Poor**: The translation is barely comprehensible, contains significant errors, and may not convey the original message. It may require extensive editing or retranslation.

    **Fair**: The translation is understandable but contains noticeable errors, inaccuracies, or awkward phrasing. It may require some editing to improve clarity and accuracy.

    **Good**: The translation is generally accurate and clear, but may contain minor errors or slight inaccuracies. It is suitable for general use but may not be perfect for critical or high-stakes applications.

    **Very Good**: The translation is highly accurate, clear, and nuanced, with only minor imperfections. It is suitable for most professional purposes and demonstrates a strong understanding of the source text.

    **Excellent**: The translation is virtually flawless, conveying the exact meaning, tone, and nuance of the original text. It is suitable for high-stakes applications, such as official publications or critical communications.

    First, explain to yourself in one sentence the reason for your rating. Then, end your response with `Rating: <rating>`.
    \end{minted}
    \caption{System prompt used to qualify the correctness of the translated sentence.}
    \label{lst:system-prompt-multi}
\end{listing}

These techniques for evaluating translation quality with an LLM are part of a larger strategy to use LLMs as judges for various tasks \cite{bavaresco_llms_2024}. Assessment in English ensures that we get the best performance from the LLM, as it excels the most in that language \cite{li_quantifying_2024}.

While the automated nature of the evaluation may result in LLM's ratings not always matching human judgments, this method offers a systematic and consistent way to assess translation quality across different languages, allowing for comparative analysis of LLM performance. To evaluate how well the LLM's ratings correspond with various prompting strategies and to test the robustness of the LLM-based evaluation, we performed a concordance analysis. This analysis sheds light on the LLM's performance and consistency in rating translation quality and identifies potential discrepancies in its evaluations.

\subsection{Comparison of GPT-4 and GPT-4o in Tokenization and Translation}

OpenAI recently introduced GPT-4o, claiming enhanced multilingual capabilities. They state, ``GPT-4o has the best vision and performance across non-English languages of any of our models" \cite{openai_models_2024}. Furthermore, improvements in tokenization compression have been claimed \cite{openai_hello_2024}. We evaluated GPT-4o against GPT-4 in tokenization and translation tasks to verify these enhancements.

We use the same sentences for both LLM versions to ensure a fair comparison. The performance of GPT-4 and GPT-4o is evaluated using the previously described methodology. We compare the ratings given by the models to assess translation quality improvements and use their respective tokenizers to evaluate enhancements in tokenization.

\section{Results}

Our evaluation of the performance differences of LLMs across various languages, using English as a reference point, focuses on these main areas: (i) how fragmentation costs in tokenization premium relate to the economic well-being of speakers, (ii) the estimated number of people affected by the disparity, (iii) the variability in LLM performance across different languages, and (iv) assessing the progress made in LLMs.

\subsection{``Poor Languages" Pay More}

A key aim of the paper is to examine and measure the link between the tokenization premium described in Eq.~\ref{eq:fragmentation} and the population-weighted average wealth across different languages formalized in Eq.~\ref{eq:wealth-indicator}. Gaining this understanding helps shed light on the economic inequalities in LLM access and the effect of tokenization on individuals from diverse socio-economic conditions.

\begin{figure}[t]
    \begin{center}
        \centerline{\includegraphics[width=\columnwidth]{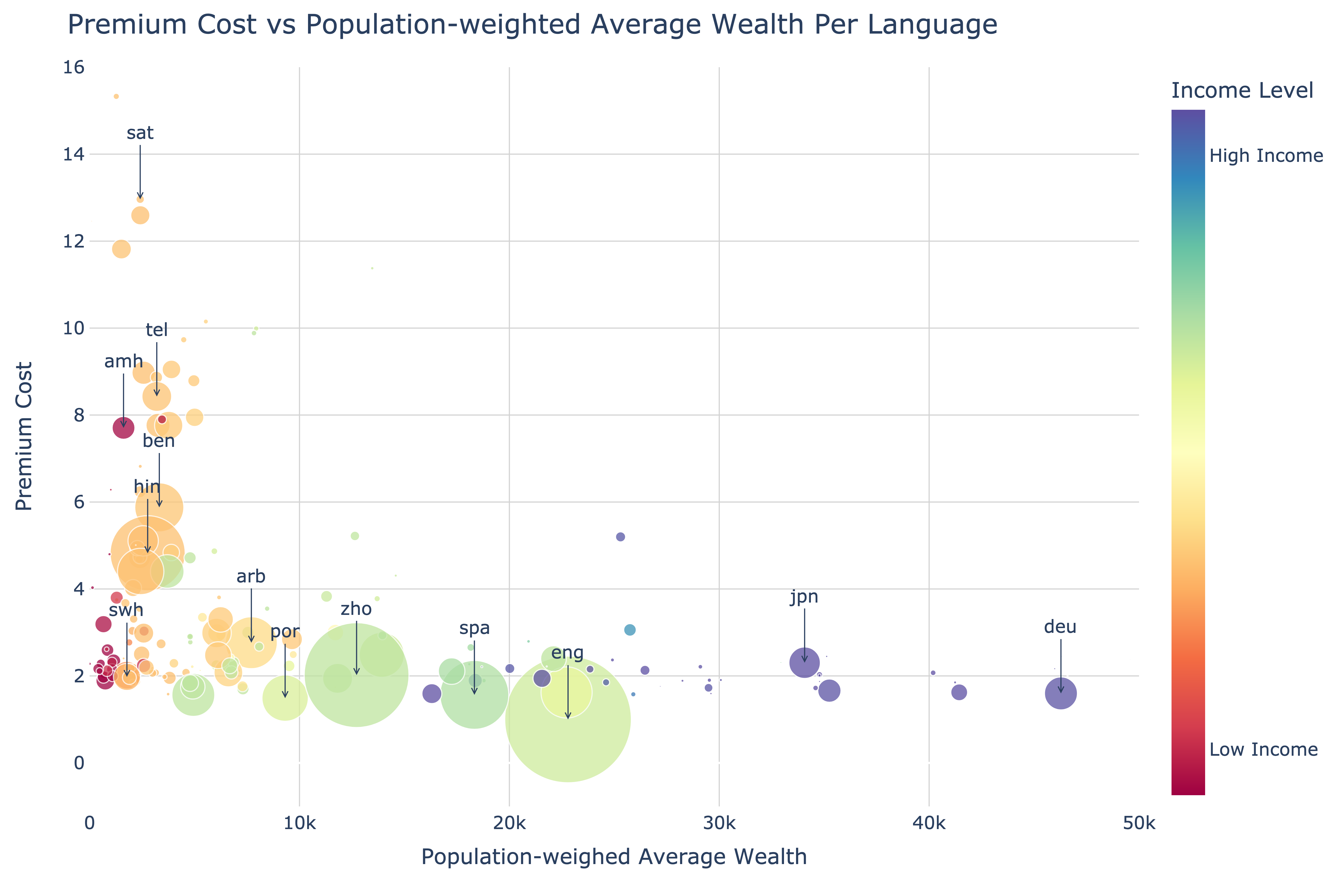}}
        \caption{The figure reveals that speakers of languages predominantly found in low- and lower-middle-income countries incur higher tokenizer costs relative to token count---a premium cost for using the GPT-4 tokenizer. The color gradient indicates the population-weighted income level associated with each language.}

        \label{fig:lang-premium-cost-50K}
    \end{center}
\end{figure}

\begin{figure}[t]
    \begin{center}
        \centerline{\includegraphics[width=\columnwidth]{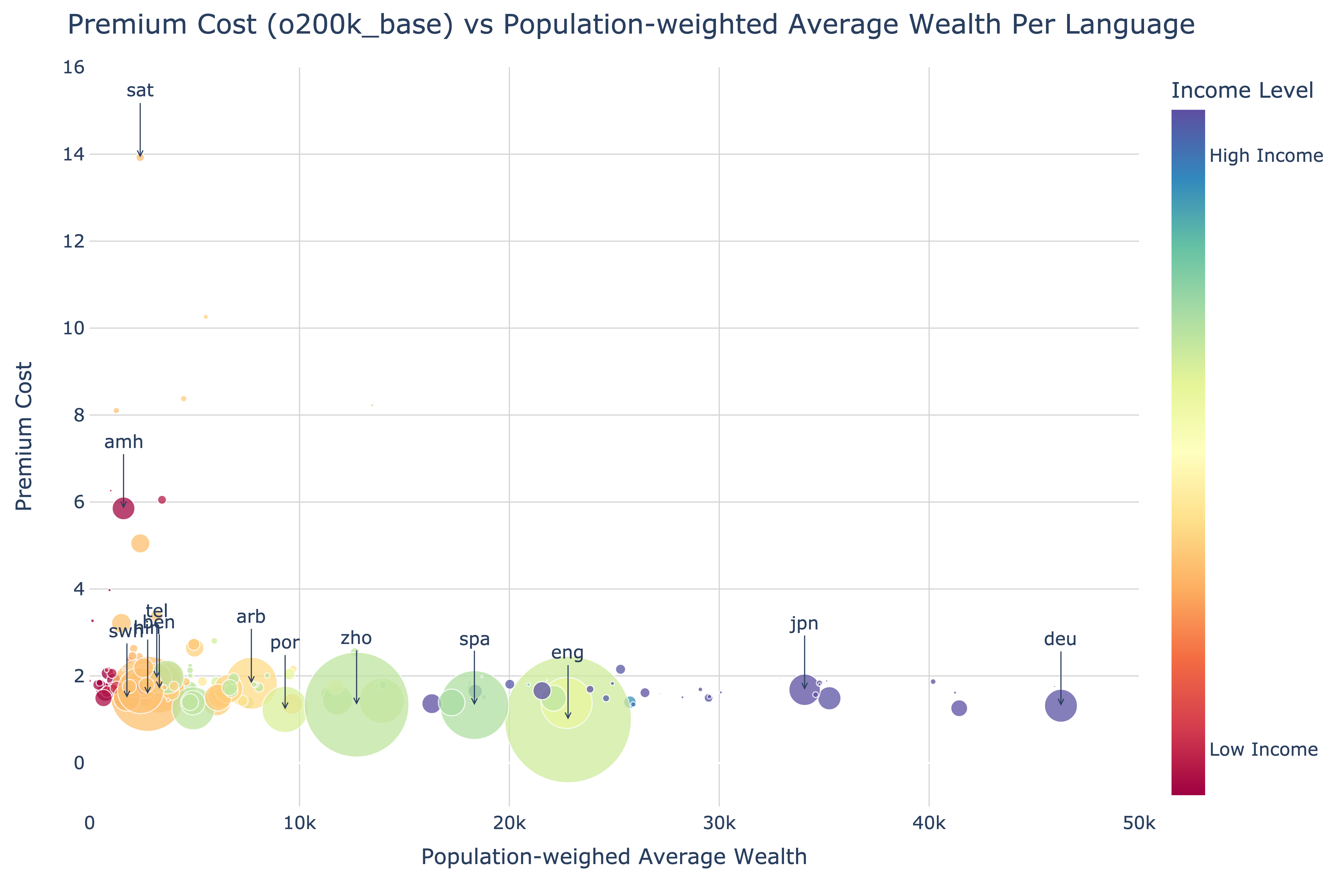}}
        \caption{The figure reveals that speakers of languages predominantly spoken in low- and lower-middle-income countries incur higher tokenizer costs relative to token count---a premium cost for using the GPT-4o tokenizer. The color gradient indicates the population-weighted income level associated with each language. However, a significant reduction in tokenization premium is observed, with most languages having premiums below 4, compared to the results produced by the GPT-4 tokenizer.}

        \label{fig:lang-premium-cost-50K-o200k}
    \end{center}
\end{figure}

Figures~\ref{fig:lang-premium-cost-50K} and \ref{fig:lang-premium-cost-50K-o200k} depict these relationships for different LLM tokenizers, presenting the results for the GPT-4 and GPT-4o tokenizers, respectively. The figures also represent additional dimensions related to the distribution of income level of countries speaking the languages as defined by Eq.~\ref{eq:income-level} and the number of speakers. These visualizations reveal several important insights.

\paragraph{Premium Costs for Non-English Languages}

Languages spoken mainly in low- and lower-middle-income countries face higher tokenization costs, shown by the elevated premium on the y-axis in both figures. Languages like Santali (sat), Telugu (tel), Amharic (amh), Bengali (ben), and Hindi (hin) have significant premium costs, ranging from 6 to 14 in Fig.~\ref{fig:lang-premium-cost-50K} and 6 to 10 in Fig.~\ref{fig:lang-premium-cost-50K-o200k}. This indicates that users of these languages incur more expenses due to increased fragmentation during tokenization---in essence, \textbf{``poor languages" pay more}.

Notably, the GPT-4o tokenizer often reduces premium costs. This improvement indicates better tokenization efficiency, backing OpenAI's claims and making it more accessible and cost-effective for users compared to GPT-4-like tokenizers.

\paragraph{Economic Disparities}

Also highlighted are the economic disparities between languages, with languages spoken in wealthier regions---such as German, Japanese, and English---positioned to the right, indicating higher population-weighted average wealth. In contrast, languages spoken in economically disadvantaged regions, such as Bengali, Amharic, and Santali, appear on the left, reflecting lower average wealth. The distribution of countries by income level represented by the color gradient reinforces these economic disparities: warmer colors indicate higher premium costs for languages mostly spoken in low-income countries, while cooler colors represent lower premium costs for languages mostly spoken in wealthier regions. This clear economic stratification underscores the affordability challenges posed by tokenization, particularly in low-income regions, where LLM services are both less accessible and more expensive.

\paragraph{Population Impact}

The size of the bubbles represents the speaker population for each language, illustrating the varying population sizes affected by tokenization costs. Smaller populations, such as those of Santali, Telugu, and Amharic, face premium costs that are up to eight times higher than those for English, despite having fewer speakers. This situation highlights the disproportionate economic burden placed on speakers of low-resource languages.

Conversely, languages with large populations, such as Chinese and Hindi, also face premium costs that are at least double those of English, yet their larger populations suggest that a significant number of users are impacted by these costs. The combination of higher tokenization premiums and larger affected populations further intensifies the economic strain for these languages, emphasizing the need for more equitable tokenization strategies to reduce costs across all languages, particularly those with higher speaker counts and lower-income regions.

\subsection{A Lower-Middle Income Trap in LLMs?}

\begin{figure}
\begin{center}
\centerline{\includegraphics[width=\columnwidth]{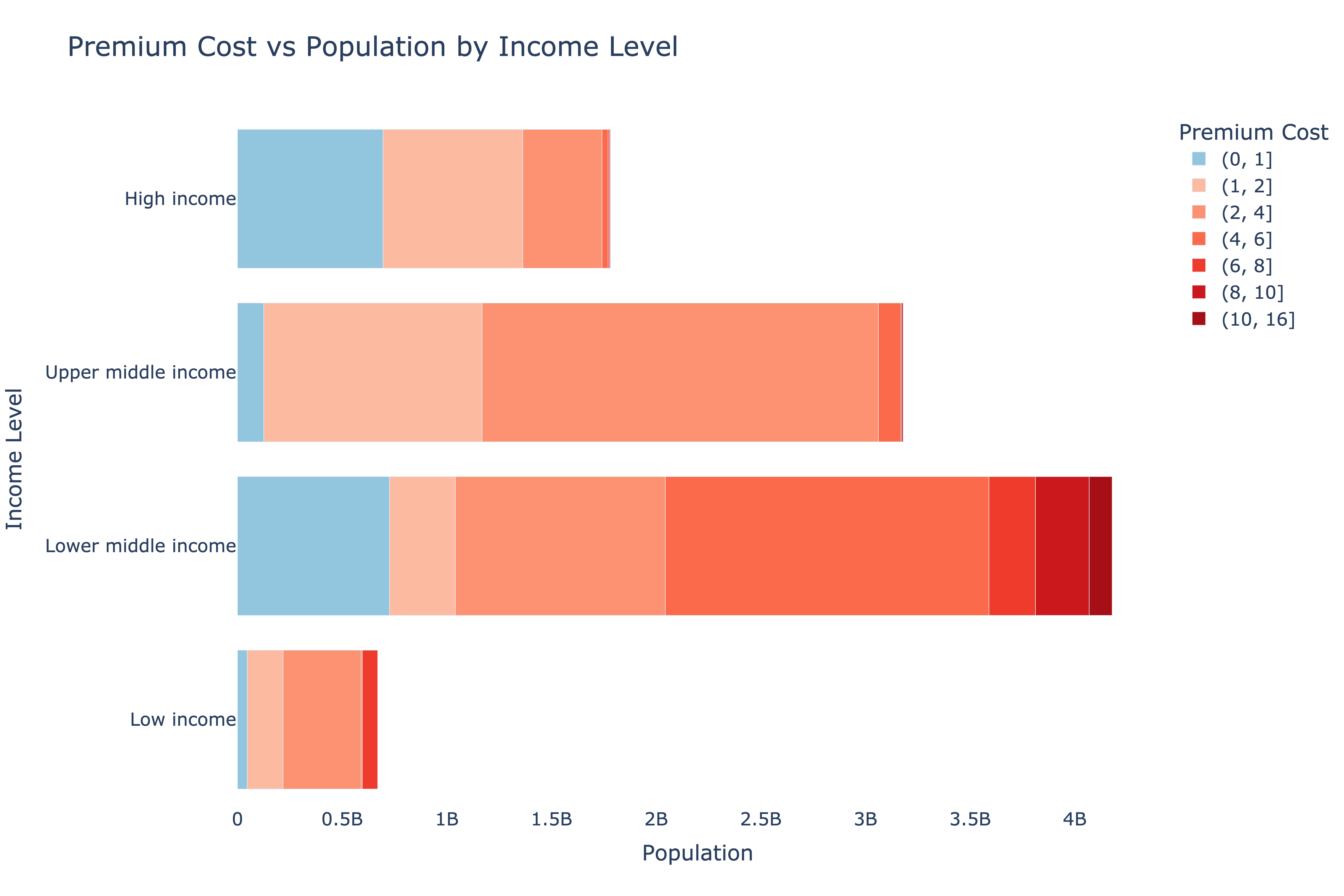}}
\caption{Stacked bar plot illustrating the population affected by each premium cost category using GPT-4. Each bar segment represents a different cost premium category, with the length indicating the proportion of the population impacted. The figure reveals that speakers of languages predominantly found in low- and lower-middle-income countries incur higher tokenizer costs relative to token count. About 1.5 billion speakers of languages in lower-middle-income countries face a premium cost between 4 to 6 times higher than that of English.}
\label{img:pop-left-behind-gpt-4}
\end{center}
\end{figure}

\begin{figure}
\begin{center}
\centerline{\includegraphics[width=\columnwidth]{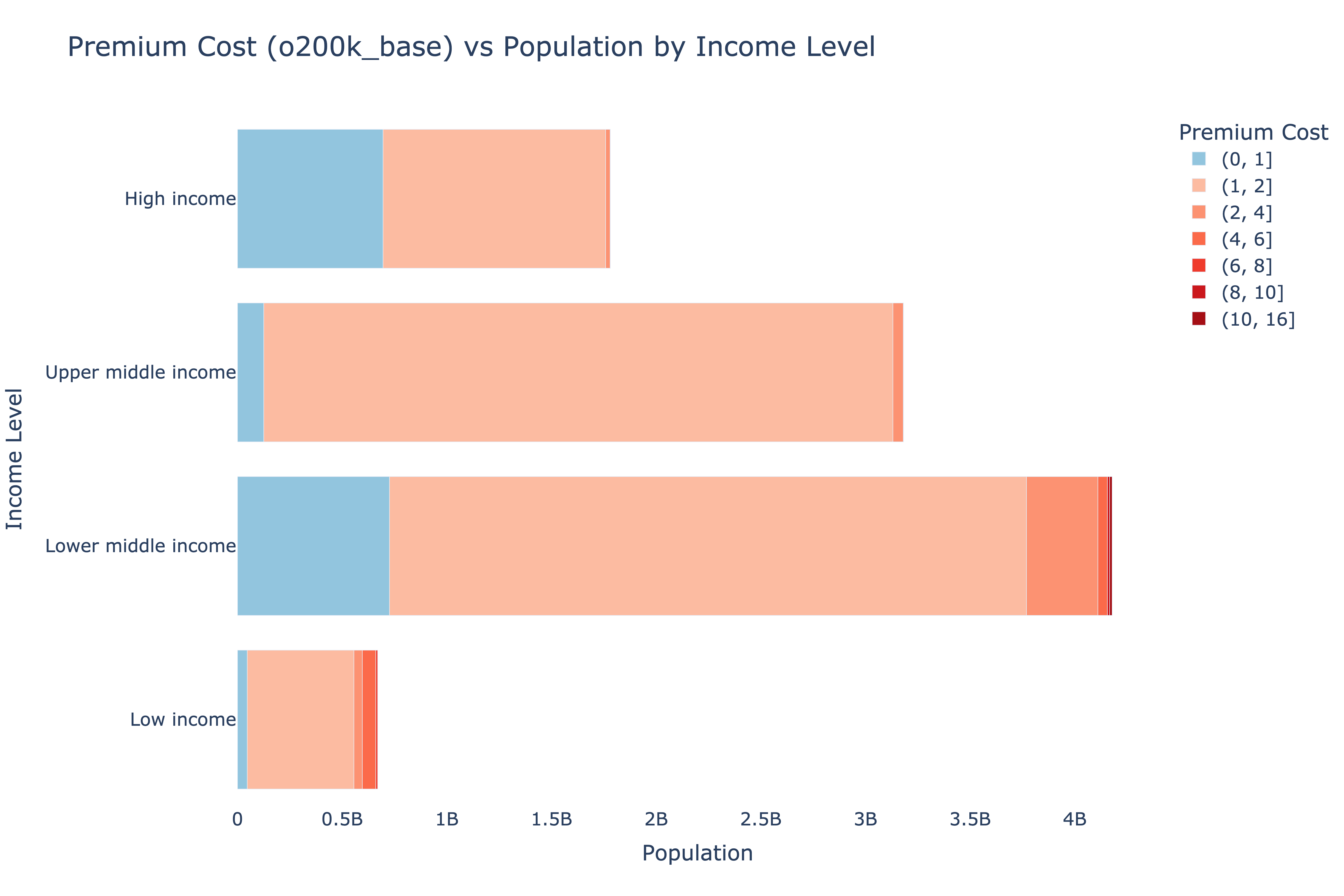}}
\caption{Stacked bar plot illustrating the population affected by each premium cost category using GPT-4o. Each bar segment represents a different cost premium category, with the length indicating the proportion of the population impacted. The figure reveals that speakers of languages predominantly found in low- and lower-middle-income countries incur higher tokenizer costs as a function of token count. However, the distribution shows an improvement over the one produced using GPT-4.}
\label{img:pop-left-behind-gpt-4o}
\end{center}
\end{figure}

With data on countries where specific languages are spoken, we can evaluate how the tokenization premium is distributed across different income levels and estimate the number of speakers impacted. Figures~\ref{img:pop-left-behind-gpt-4} and \ref{img:pop-left-behind-gpt-4o} illustrate these distributions for the GPT-4 and GPT-4o models, respectively. In both instances, we observe that speakers of lower-middle-income languages are not only the largest group affected by the tokenization premium but also incur some of the highest premiums.

\subsubsection{GPT-4 Tokenizer: The Population Impacted}

The GPT-4 model tokenizer results show that high-income countries face much lower tokenization costs. In these regions, about 38.89\% of people speak English, and 37.51\% have premiums ranging from 1 to 2 times. A very small fraction (around 0.02\%) faces premiums of 8 to 10 times, with an even smaller group incurring 10 to 16 times the cost.

Conversely, the upper-middle-income demographic confronts a moderately higher overall tokenization cost burden. Although many individuals in this group still encounter low premiums (approximately 3.89\% English speakers, and 31.81\% with premiums between 1 and 2), a significantly larger portion--around 60.82\%--faces premiums ranging from 2 to 4 times. This indicates that while some within this income bracket experience costs comparable to those in high-income nations, the majority bear increased tokenization burdens. The segment affected by premiums of 6 to 8 times remains negligible at 0.07\%, and similarly, just 0.07\% deal with a premium of 10 to 16 times.

In the lower-middle-income group, tokenization premiums show a concerning trend. Only 17.51\% face a premium of 0 to 1, while 7.79\% see between 1 and 2. The majority, 36.03\%, experience premiums of 4 to 6, and 13.63\% have premiums of at least 6 times the English cost, with 5.16\% in the 6 to 8 range and 2.55\% facing premiums of 10 to 16 times. This suggests that speakers in lower-middle-income countries are more heavily affected by higher tokenization premiums than those in upper-middle- and high-income countries.

In low-income countries, a significant portion of the population (56.92\%) experiences premium costs falling within the 2 to 4 times category. This represents the greatest concentration of premiums for this income bracket, indicating that many individuals are dealing with moderate tokenization expenses. Additionally, 24.51\% of the populace encounters premiums ranging from 1 to 2 times, and 6.35\% speak English. On the other hand, 11.49\% face premiums between 6 and 8 times, highlighting the uneven burden on people in these areas. Importantly, the analysis reveals no languages in low-income countries with premiums above 8 times, which contrasts with lower-middle-income countries where around 2.55\% of the population faces premiums as high as 16 times.

In general, speakers in low- and lower-middle-income countries face much higher tokenization premiums, often more than four times that of English speakers. In contrast, those in high- and upper-middle-income countries usually see premiums of 2 times or less. To ensure linguistic inclusivity in GPT-4 LLMs, addressing these tokenization cost differences is crucial to avoid exclusion and unequal access for speakers of less common languages in poorer regions.

\subsubsection{GPT-4o Tokenizer: The Population Impacted}

In contrast, the data using the GPT-4o tokenizer appear promising. In high-income groups, about 38.89\% of the population faces a premium of 0 to 1 time the English tokenization cost, and approximately 59.83\% fall within the 1 to 2 times range. Only 1.19\% face premiums between 2 and 4 times, and less than 0.01\% experience premiums between 4 and 16 times. This aligns with previous results from the GPT-4 tokenizer, showing that high-income countries consistently have lower tokenization premiums.

In the upper-middle-income group, about 94.56\% of people face low premiums (1 to 2 times), with only 1.51\% encountering premiums between 2 and 4 times, and very few facing higher premiums. This shows a slight change from the GPT-4 tokenizer results where this group had more distribution in the 2 to 4 premium range. However, tokenization burdens are still relatively low for this income group under both tokenizers.

Conversely, the lower-middle-income group has a more diverse premium distribution, with around 72.86\% facing 1 to 2 times premiums and 8.07\% encountering 2 to 4 times premiums.

Only about 1.04\% of the population faces premiums between 4 and 6 times, and just around 0.5\% faces premiums greater than 6. This marks a notable shift from the GPT-4 tokenizer, with fewer individuals encountering premiums over 4 times that of English.

It is encouraging to see a significant shift in premium distribution for speakers in low-income countries, with about 76.73\% now falling within the lower range of 1 to 2 times. This is an improvement from the GPT-4 tokenizer, where 56.92\% faced premiums of 2 to 4 times. Similarly, the proportion of speakers dealing with premiums between 4 to 6 times has dropped to 9.69\%, compared to 11.49\% experiencing 6 to 8 times with the GPT-4 tokenizer. This indicates a notable reduction in tokenization costs for these regions with the GPT-4o tokenizer.

\subsection{LLM Performance Across Languages}

While we have explored the economic effects of LLMs due to varying tokenization costs among languages, it is equally vital to consider their performance across different languages. Paying higher costs could be justified if the model consistently delivers high-quality results in all languages. To explore this matter, we evaluate the performance of LLMs across a varied sample of languages to determine if there are performance gaps that may exacerbate the disadvantages for speakers of languages with higher costs. Identifying such disparities is essential, as poor model performance, along with increased tokenization expenses, would further increase the inequities experienced by these language communities.

\begin{table}[ht]
    \centering
    \caption{Comparison of back-translation accuracy between GPT-4 Turbo and GPT-4o models across various languages, evaluated by an LLM judge. The first section presents results where the LLM judge provided a direct binary rating (zero-shot), while the second section reflects ratings given after the LLM judge was prompted to explain its reasoning before making a final judgment (chain-of-thought). All values are expressed as percentages.}
    \label{tab:back-translate-binary}  
        \begin{tabular}{llcccc}
            \toprule
            \textbf{Language} & \textbf{Code (ISO 639)} & \multicolumn{2}{c}{\textbf{GPT-4 Turbo}} & \multicolumn{2}{c}{\textbf{GPT-4o}}                                         \\
                              &                         & \textbf{Incorrect}                       & \textbf{Correct}                    & \textbf{Incorrect} & \textbf{Correct} \\
            \midrule
            \multicolumn{6}{c}{\textbf{Rating without explanation} (zero-shot prompting)}                                                                                                              \\
            \midrule
            \multicolumn{6}{l}{\textbf{Low-income languages}}                                              \\                                                                                              
            Dzongkha          & dzo                     & 100.00                                   & -                                   & 98.50              & 1.50             \\
            Tamasheq          & taq                     & 99.20                                    & 0.80                                & 98.80              & 1.20             \\
            Kabiyè            & kbp                     & 99.70                                    & 0.30                                & 99.20              & 0.80             \\
            Nuer              & nus                     & 99.90                                    & 0.10                                & 99.40              & 0.60             \\
            \midrule
        \multicolumn{6}{l}{\textbf{Low-middle-income languages}}                                                                                                                    \\
            Shan              & shn                     & 99.20                                    & 0.80                                & 98.80              & 1.20             \\
            Santhali          & sat                     & 100.00                                   & -                                   & 99.90              & 0.10             \\
            Odia              & ory                     & 54.86                                    & 45.14                               & 42.03              & 57.97            \\
            Hindi             & hin                     & 43.43                                    & 56.57                               & 39.22              & 60.78            \\
            Bengali           & ben                     & 52.66                                    & 47.34                               & 44.83              & 55.17            \\
            Urdu              & urd                     & 47.94                                    & 52.06                               & 44.33              & 55.67            \\
            \midrule
        \multicolumn{6}{l}{\textbf{High-population languages}}                                                                                                                                                   \\
            Chinese           & zho                     & 34.30                                    & 65.70                               & 30.49              & 69.51            \\
            Spanish           & spa                     & 19.86                                    & 80.14                               & 18.66              & 81.34            \\
            Standard Arabic   & arb                     & 27.88                                    & 72.12                               & 23.47              & 76.53            \\
            French            & fra                     & 16.45                                    & 83.55                               & 15.35              & 84.65            \\
            \midrule
            \multicolumn{6}{c}{\textbf{Rating with explanation} (chain-of-thought prompting)}                                                                                                                 \\
            \midrule
            \multicolumn{6}{l}{\textbf{Low-income languages}}                                              \\                                                                                              
            
            Dzongkha          & dzo                     & 99.90                                    & 0.10                                & 97.10              & 2.90             \\
            Tamasheq          & taq                     & 98.80                                    & 1.20                                & 97.10              & 2.90             \\
            Kabiyè            & kbp                     & 99.20                                    & 0.80                                & 98.50              & 1.50             \\
            Nuer              & nus                     & 99.30                                    & 0.70                                & 99.20              & 0.80             \\
            \midrule
        \multicolumn{6}{l}{\textbf{Low-middle-income languages}}                                                                                                                    \\
            Shan              & shn                     & 98.40                                    & 1.60                                & 97.90              & 2.10             \\
            Santhali          & sat                     & 100.00                                   & -                                   & 99.90              & 0.10             \\
            Odia              & ory                     & 46.64                                    & 53.36                               & 35.01              & 64.99            \\
            Hindi             & hin                     & 37.21                                    & 62.79                               & 32.90              & 67.10            \\
            Bengali           & ben                     & 41.42                                    & 58.58                               & 36.21              & 63.79            \\
            Urdu              & urd                     & 39.02                                    & 60.98                               & 35.11              & 64.89            \\
            \midrule
        \multicolumn{6}{l}{\textbf{High-population languages}}                                                                                                                                                   \\
            Chinese           & zho                     & 22.07                                    & 77.93                               & 20.56              & 79.44            \\
            Spanish           & spa                     & 13.74                                    & 86.26                               & 12.04              & 87.96            \\
            Standard Arabic   & arb                     & 22.07                                    & 77.93                               & 18.66              & 81.34            \\
            French            & fra                     & 11.74                                    & 88.26                               & 10.43              & 89.57            \\
            \bottomrule
        \end{tabular}
\end{table}

We utilize the translation task as a proxy for assessing LLM performance, as outlined in Section~\ref{sec:translation-proxy-llm}. The selection of languages is based on the following criteria: (i) the top 3 languages with the highest premiums from low-income countries, (ii) the top 3 languages by total population with at least a 4x premium in low-income countries, (iii) the top 3 languages with the highest premiums from lower-middle-income countries, (iv) the top 3 languages by total population with at least a 4x premium in lower-middle-income countries, and (v) the top 5 languages by total global population. This selection process aims to identify languages that are particularly disadvantaged due to a combination of high tokenization costs, economic factors, and large speaker populations, ensuring that our focus is on the most affected groups. We end up with 14 languages: Dzongkha (dzo), Tamasheq (taq), Kebiy`e (kbp), Nuer (nus), Shan (shn), Santali (sat), Odia (ory), Hindi (hin), Bengali (ben), Urdu (urd), Chinese (zho), Spanish (spa), Arabic (arb), and French (fra).

We chose the GPT-4o model as the LLM judge because it performs best among current LLMs \cite{openai_hello_2024}. Both the translation and evaluation tasks are conducted through separate calls to the model, each guided by specific system prompts. Table~\ref{tab:back-translate-binary} presents the results of our binary assessment of translation quality, while Table~\ref{tab:back-translate-five-point} presents the results of the five-point scale assessment.

As anticipated, the results indicate that languages linked to low-income regions perform poorly in back-translation tasks, and many sentences are incorrectly translated. Dzongkha and Nuer had almost 0\% accuracy. GPT-4 Turbo showed a slight improvement over GPT-4o, with Tamasheq translations increasing from 0.80\% to 1.20\%. However, these LLMs still prove inadequate for accurately translating low-income languages, limiting their practical use for these linguistic groups.

For languages from lower-middle-income countries, outcomes vary. Shan and Santhali perform very poorly, with nearly all translations incorrect. Santhali, in particular, has dismal results with almost no accurate translations under GPT-4o (0.10\%) and none under GPT-4 Turbo. Conversely, Odia, Hindi, Bengali, and Urdu do better, sometimes achieving around 50\% accuracy. The GPT-4o model shows improvement; for instance, Odia's accuracy rose from 45.14\% with GPT-4 Turbo to 57.97\%. Hindi and Bengali also saw gains, with accuracies reaching 60.78\% and 55.17\%.

High-population languages like Chinese and Spanish perform relatively well, with over 65\% correct translations for GPT-4o. However, even widely spoken languages show errors; Chinese still has more than 30\% incorrect translations in both models. Generally, GPT-4o outperforms GPT-4 Turbo, improving Spanish accuracy from 80.14\% to 81.34\% and French from 83.55\% to 84.65\%. Despite higher accuracy, there is still room for improvement in translation reliability.

The influence of chain-of-thought prompting in the second section of the table is particularly significant. Although overall accuracy remained relatively stable for low-income and lower-middle-income languages, this method appears to enhance the detection of correct translations in several cases slightly. For example, Dzongkha's accuracy rose from 1.50\% with zero-shot prompting to 2.90\% with chain-of-thought prompting in GPT-4o. Similar improvements were observed in Tamasheq, Kabiyè, and Nuer, despite these languages continuing to experience high error rates. The effect of chain-of-thought prompting is more pronounced in high-population languages, leading to substantial assessment improvements, such as in Chinese (from 65.70\% to 77.93\% with GPT-4 Turbo and from 69.51\% to 79.44\% with GPT-4o) and French (from 83.55\% to 88.26\% with GPT-4 Turbo and from 84.65\% to 89.57\% with GPT-4o).

Our findings suggest that GPT-4o slightly outperforms GPT-4 Turbo, particularly in lower-middle-income and populous languages. However, persistent translation quality issues remain for lower-income languages, where the performance gains are minimal. These results highlight the ongoing challenges in LLM performance for underrepresented linguistic communities. While chain-of-thought prompting shows an increased rate of translations being assessed as correct, its effectiveness varies significantly between languages. Additionally, our concordance analysis, as shown in Table~\ref{tab:back-translate-judge-concordance}, indicates that chain-of-thought prompting (utilizing explanations) more effectively classifies translations as ``Correct" when rated ``Excellent" or ``Very Good" by a model tasked with assessing quality based on a rubric presented in Listing~\ref{lst:system-prompt-multi}, compared to the zero-shot method. Furthermore, chain-of-thought prompting appears to better differentiate translation quality from the zero-shot model when translations are rated only as ``Good."

\begin{table}[ht]
    \centering    
    \caption{Comparison of back-translation accuracy between GPT-4 Turbo and GPT-4o models across various languages, evaluated by an LLM judge. All values are expressed as percentages; empty cells are represented by `-'.}
    \label{tab:back-translate-five-point}
        \begin{tabular}{llccccc}
            \toprule
            \textbf{Language} & \textbf{Code (ISO 639)} & \textbf{Poor} & \textbf{Fair} & \textbf{Good} & \textbf{Very Good} & \textbf{Excellent} \\
            \midrule
            \multicolumn{7}{c}{\textbf{GPT-4 Turbo}}                                                                                              \\
            \midrule
            \multicolumn{7}{l}{\textbf{Low-income languages}}                                              \\
            Dzongkha          & dzo                     & 97.89         & 1.71          & 0.40          & -                  & -                  \\
            Tamasheq          & taq                     & 89.07         & 5.42          & 4.41          & 0.60               & 0.50               \\
            Kabiyè            & kbp                     & 92.78         & 3.61          & 2.91          & 0.40               & 0.30               \\
            Nuer              & nus                     & 94.38         & 3.11          & 2.31          & 0.20               & -                  \\
            \midrule
            \multicolumn{7}{l}{\textbf{Low-middle-income languages}}                                              \\
            Shan              & shn                     & 89.47         & 4.91          & 4.21          & 1.00               & 0.40               \\
            Santhali          & sat                     & 99.90         & 0.10          & -             & -                  & -                  \\
            Odia              & ory                     & 9.63          & 11.03         & 41.62         & 26.58              & 11.13              \\
            Hindi             & hin                     & 3.61          & 8.33          & 42.23         & 34.30              & 11.53              \\
            Bengali           & ben                     & 4.81          & 10.53         & 44.83         & 27.78              & 12.04              \\
            Urdu              & urd                     & 4.61          & 9.33          & 41.83         & 31.59              & 12.64              \\
            \midrule
            \multicolumn{7}{l}{\textbf{High-population languages}}                                              \\
            Chinese           & zho                     & 1.10          & 3.01          & 32.10         & 45.84              & 17.95              \\
            Spanish           & spa                     & 0.70          & 2.21          & 27.58         & 53.96              & 15.55              \\
            Standard Arabic   & arb                     & 2.01          & 4.71          & 30.59         & 49.95              & 12.74              \\
            French            & fra                     & 0.70          & 1.60          & 18.36         & 66.30              & 13.04              \\
            \midrule
            \multicolumn{7}{c}{\textbf{GPT-4o}}                                                                                                   \\
            \midrule
            \multicolumn{7}{l}{\textbf{Low-income languages}}                                              \\
            Dzongkha          & dzo                     & 78.74         & 11.94         & 7.52          & 1.10               & 0.70               \\
            Tamasheq          & taq                     & 84.55         & 7.72          & 5.32          & 1.71               & 0.70               \\
            Kabiye            & kbp                     & 89.47         & 5.32          & 4.01          & 0.60               & 0.60               \\
            Nuer              & nus                     & 93.98         & 2.61          & 2.91          & 0.30               & 0.20               \\
            \midrule
            \multicolumn{7}{l}{\textbf{Low-middle-income languages}}                                              \\
            Shan              & shn                     & 87.96         & 5.92          & 4.21          & 1.30               & 0.60               \\
            Santhali          & sat                     & 99.10         & 0.60          & 0.30          & -                  & -                  \\
            Odia              & ory                     & 5.42          & 8.93          & 37.11         & 35.11              & 13.44              \\
            Hindi             & hin                     & 3.21          & 6.32          & 39.22         & 36.71              & 14.54              \\
            Bengali           & ben                     & 3.71          & 9.23          & 40.52         & 34.20              & 12.34              \\
            Urdu              & urd                     & 4.01          & 8.33          & 39.32         & 35.91              & 12.44              \\
            \midrule
            \multicolumn{7}{l}{\textbf{High-population languages}}                                              \\
            Chinese           & zho                     & 0.80          & 2.51          & 28.69         & 49.95              & 18.05              \\
            Spanish           & spa                     & 0.20          & 2.51          & 26.38         & 54.96              & 15.95              \\
            Standard Arabic   & arb                     & 1.10          & 2.31          & 27.78         & 54.86              & 13.94              \\
            French            & fra                     & 0.60          & 1.60          & 16.05         & 66.70              & 15.05              \\
            \bottomrule
        \end{tabular}
\end{table}

\begin{table}[h]
    \caption{Concordance analysis of the different methods of automated assessment of the back translation quality}\label{tab:back-translate-judge-concordance}%
    \centering
    \begin{tabular}{@{}lcccc@{}}
        \toprule
        \multicolumn{5}{c}{\textbf{GPT-4o}}                                                                                                          \\
        \midrule
        \multirow{2}{*}{Rating} & \multicolumn{2}{c}{Binary without explanation} & \multicolumn{2}{c}{Binary with explanation}                       \\
        \cmidrule(lr){2-3} \cmidrule(lr){4-5}
                                & Incorrect                                      & Correct                                     & Incorrect & Correct \\
        \midrule
        Excellent               & 3.64                                           & 96.36                                       & 1.27      & 98.73   \\
        Very Good               & 6.63                                           & 93.37                                       & 2.66      & 97.34   \\
        Good                    & 69.73                                          & 30.27                                       & 54.04     & 45.96   \\
        Fair                    & 99.87                                          & 0.13                                        & 99.34     & 0.66    \\
        Poor                    & 99.98                                          & 0.02                                        & 100.00     & 0.00    \\
        \midrule
        \multicolumn{5}{c}{\textbf{GPT-4 Turbo}}                                                                                                     \\
        \midrule
        \multirow{2}{*}{Rating} & \multicolumn{2}{c}{Binary without explanation} & \multicolumn{2}{c}{Binary with explanation}                       \\
        \cmidrule(lr){2-3} \cmidrule(lr){4-5}
                                & Incorrect                                      & Correct                                     & Incorrect & Correct \\
        \midrule
        Excellent               & 3.91                                           & 96.09                                       & 1.30      & 98.70   \\
        Very Good               & 6.52                                           & 93.48                                       & 2.46      & 97.54   \\
        Good                    & 71.18                                          & 28.82                                       & 54.50     & 45.50   \\
        Fair                    & 100.00                                          & 0.00                                        & 99.42     & 0.58    \\
        Poor                    & 100.00                                          & 0.00                                        & 100.00     & 0.00    \\
        \botrule
    \end{tabular}
\end{table}

\subsection{Improvements in Premium Costs and Performance}

\begin{figure}[t]
    \begin{center}
        \centerline{\includegraphics[width=\columnwidth]{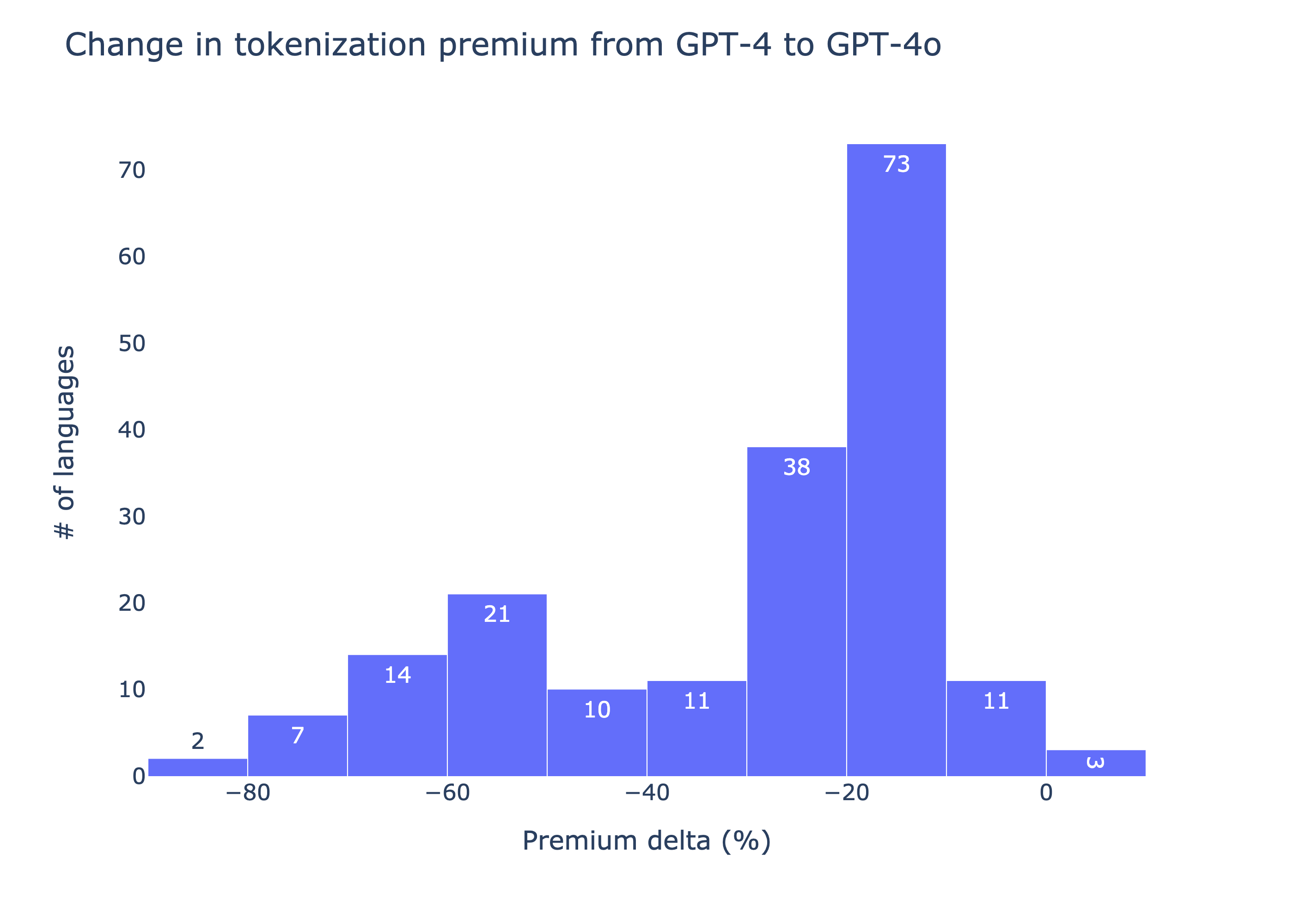}}
        \caption{Distribution of changes in tokenization premium from GPT-4 to GPT-4o across various languages. The x-axis represents the percentage change in premium, while the y-axis indicates the number of languages that fall into each change category. The histogram reveals that the most common tokenization premium reduction affecting 73 languages is around -20\%. Other significant reductions include -60\% and -40\%, impacting 21 and 38 languages, respectively. Only a small number of languages saw an increase in tokenization premium, namely Santhali (7.44\%) and Tamazight (1.06\%).}

        \label{fig:premium-change}
    \end{center}
\end{figure}

\begin{figure}[t]
    \begin{center}
        \centerline{\includegraphics[width=\columnwidth]{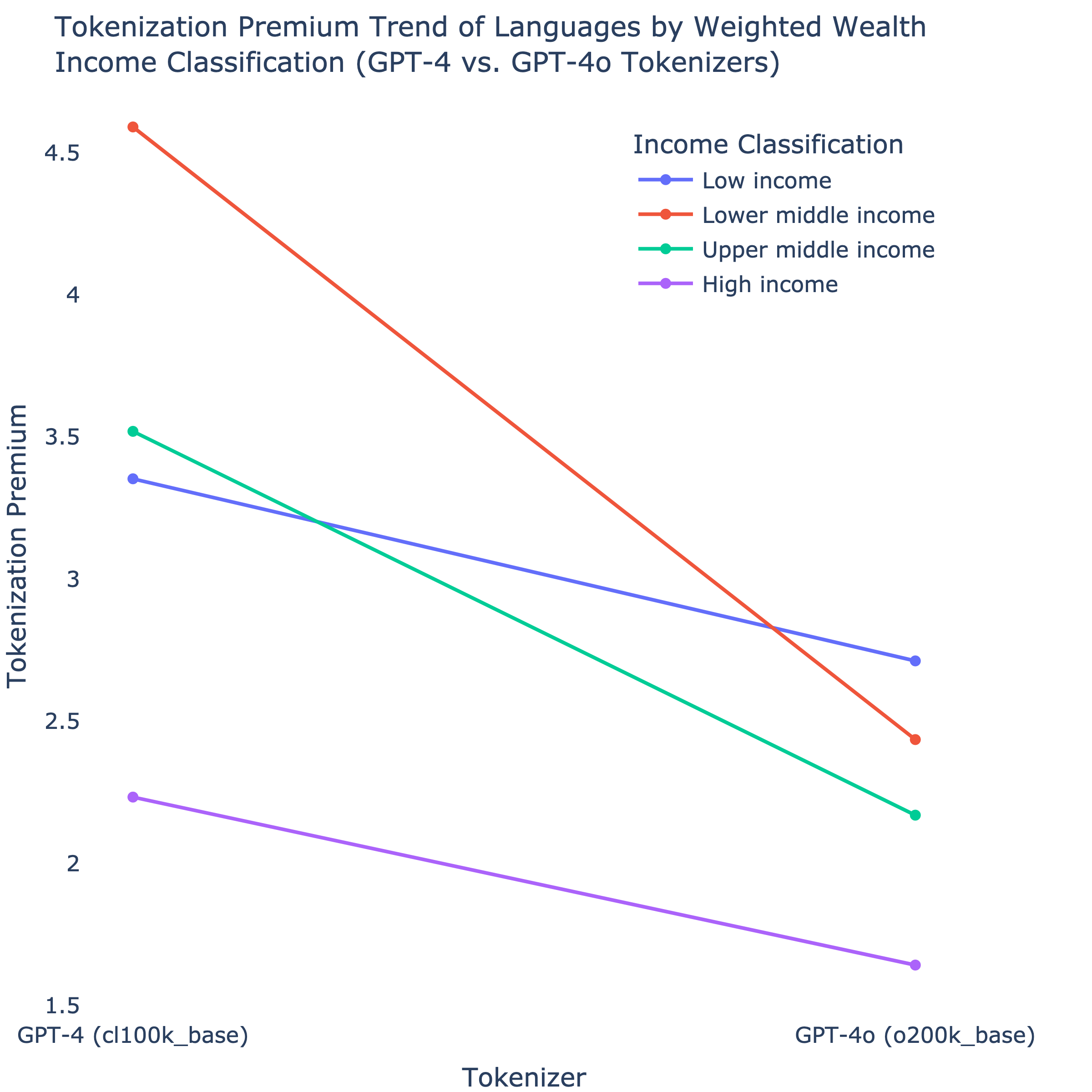}}
        \caption{This figure compares the tokenization premium across groups of languages by weighted wealth income classification, using GPT-4 (cl100k\_base) and GPT-4o (o200k\_base) tokenizers. The y-axis represents the tokenization premium, and each line connects the values based on the tokenizer used by GPT-4 to GPT-4o, demonstrating the reduction in tokenization premium. The figure highlights a more significant decrease in tokenization premiums for lower-middle-income groups.}

        \label{fig:tokenization-premium-trend}
    \end{center}
\end{figure}

Our analysis shows that tokenization premiums have decreased from GPT-4 to GPT-4o for most languages. Only Santali (sat) and Tamazight (tzm) see increases in tokenization premiums. The median reduction is 20.93\%, with the average decrease being 30.13\%. Santali's premium rose by 7.44\%, while Tamazight's increased slightly by 1.06\%. This indicates improved tokenization efficiency in GPT-4o, lowering costs for non-English languages. Fig.~\ref{fig:premium-change} and \ref{fig:tokenization-premium-trend} summarize these changes, and Table~\ref{tab:premium-cost-comparison} compares tokenization costs for select languages between GPT-4 and GPT-4o. Notably, languages with population-weighted wealth classified as lower-middle-income see a larger overall improvement in tokenization premium.

These results are promising, indicating progress in reducing tokenization cost disparities across languages. The overall decrease in premiums shows better tokenization efficiency, making LLMs more accessible and affordable for non-English speakers. Yet, the rise in premiums for some languages emphasizes the need for continued research to tackle their specific challenges.

It is also important to note that although GPT-4o shows promising improvements in tokenization premiums, its performance in low-resource languages is still lacking.

\begin{table}[h]
    \caption{Premium cost comparison between tokenizers used by GPT-4 and GPT-4o for selected languages}\label{tab:premium-cost-comparison}%
    \begin{tabular}{@{}lccccc@{}}
        \toprule
        \multirow{2}{*}{Language}                                   & \multirow{2}{*}{Code (ISO 639-3)} & \multicolumn{2}{c}{Premium Cost\footnotemark[1]} & \multirow{2}{*}{Percent Difference (\%)}            \\
        \cmidrule(lr){3-4}
                                                                    &                                   & GPT-4 (cl100k)                                   & GPT-4o (o200k)                                      \\
        \midrule
        \multicolumn{5}{l}{\textbf{Low-income languages}}                                                                                                                                                        \\
        Dzongkha\footnotemark[2]                                    & dzo                               & 12.46                                            & 9.22                                     & -25.98\% \\
        Tamasheq\footnotemark[3]\textsuperscript{,}\footnotemark[4] & taq                               & 6.28                                             & 6.26                                     & -0.34\%  \\
        Kabiyè\footnotemark[2]\textsuperscript{,}\footnotemark[3]   & kbp                               & 4.80                                             & 3.97                                     & -17.19\% \\
        Nuer\footnotemark[3]                                        & nus                               & 4.03                                             & 3.27                                     & -18.92\% \\

        \midrule
        \multicolumn{5}{l}{\textbf{Low-middle-income languages}}                                                                                                                                                 \\
        Shan\footnotemark[4]                                        & shn                               & 15.33                                            & 8.10                                     & -47.14\% \\
        Santali\footnotemark[4]                                     & sat                               & 12.96                                            & 13.93                                    & 7.44\%   \\
        Odia\footnotemark[4]                                        & ory                               & 12.59                                            & 5.05                                     & -59.91\% \\
        Hindi\footnotemark[5]\textsuperscript{,}\footnotemark[6]    & hin                               & 4.82                                             & 1.59                                     & -67.08\% \\
        Bengali\footnotemark[5]                                     & ben                               & 5.88                                             & 1.71                                     & -70.89\% \\
        Urdu\footnotemark[5]                                        & urd                               & 4.40                                             & 1.67                                     & -62.12\% \\

        \midrule
        \multicolumn{5}{l}{\textbf{High-population languages}}                                                                                                                                                   \\
        English\footnotemark[6]                                     & eng                               & 1.00                                             & 1.00                                     & 0.00\%   \\
        Chinese\footnotemark[6]                                     & zho                               & 2.02                                             & 1.34                                     & -33.64\% \\
        Spanish\footnotemark[6]                                     & spa                               & 1.57                                             & 1.33                                     & -15.00\% \\
        Arabic\footnotemark[6]                                      & arb                               & 2.76                                             & 1.83                                     & -33.71\% \\
        French\footnotemark[6]                                      & fra                               & 1.62                                             & 1.38                                     & -14.89\% \\
        \botrule
    \end{tabular}
    \footnotetext{This table shows that speakers of languages primarily prevalent in low- and middle-income countries face elevated tokenizer premium costs.}
    \footnotetext[1]{The premium cost is calculated relative to English (eng) as the baseline. The selection is based on premium using the GPT-4 Turbo tokenizer.}
    \footnotetext[2]{Top 3 languages with the highest premium among low-income weighted wealth languages.}
    \footnotetext[3]{Top 3 languages by total population with at least 4x premium among low-income weighted wealth languages.}
    \footnotetext[4]{Top 3 languages with the highest premium among low-middle-income weighted wealth languages.}
    \footnotetext[5]{Top 3 languages by total population with at least 4x premium among low-middle-income weighted wealth languages.}
    \footnotetext[6]{Top 5 languages by total population.}
\end{table}

\section{Discussion}

\subsection{The Double Jeopardy}

The findings show that users of low-resource languages face a double jeopardy with LLMs. They not only bear higher tokenization costs, which disproportionately impact lower-middle and low-income groups, but also experience poor performance in translation and processing with these models, even with efficiency gains in GPT-4o.

The data indicate that, although there have been some successes in lowering tokenization premiums, specifically with the GPT-4o model, these benefits are uneven. Lower-middle-income languages, in particular, bear the brunt of high tokenization costs, exacerbating their existing economic challenges.

While lower tokenization premiums for some languages are a step forward, the high costs for low-resource languages underscore the need for more tailored solutions. For many speakers of these languages, high tokenization fees can be especially restrictive, limiting access to advanced language technologies like LLMs.

The problem extends beyond tokenization's economic impact. Even with reduced tokenization premiums, LLMs perform poorly on tasks such as translation in low-resource languages. This leads to higher computational costs and subpar model performance for these speakers. For instance, although the GPT-4o model offers lower tokenization costs, it still struggles with accurate translations in many low-resource languages, negating any cost savings due to its less reliable outputs.

This ``double jeopardy"---higher tokenization costs combined with poorer model performance---creates a significant barrier for low-resource language speakers, limiting their ability to benefit from advancements in LLMs. It highlights the inherent inequities in current LLM systems. While high-resource languages, particularly those spoken by high-income groups, enjoy both low costs and high model performance, low-resource language speakers face both economic and functional disadvantages. This could further entrench existing divides, where only a select group of languages, typically tied to wealthier populations, reap the full benefits of cutting-edge AI technologies such as LLMs---echoing the disparities seen during the rise of digitalization.

Addressing this dual challenge requires sustained research and development to improve the proficiency and effectiveness of LLMs across diverse languages, as well as a dedicated effort to gather high-quality data for under-resourced languages. Important strategies include enhancing the availability of training data, fine-tuning models for specific linguistic traits, and developing cost-efficient tokenization algorithms. Additionally, creating policies that involve marginalized communities in AI design and deployment is crucial for ensuring equitable distribution of LLM benefits.

\subsection{Tokenization Inequities Increase the Climate Impact for Everyone}

Inefficiencies in tokenization, particularly for languages with complex morphology or those lacking robust linguistic resources, present significant social and environmental challenges. Fragmentation during tokenization inflates token counts, increasing the computational burden as shown in Equation~\ref{eq:flop}, especially for low-resource languages. These inefficiencies have tangible consequences, affecting the accessibility of language technologies and their broader environmental impact.

As discussed in Section~\ref{subsec:flops}, the number of tokens processed is directly proportional to the carbon footprint of a model. LLMs rely heavily on tokenization to segment input text. When tokenization leads to excessive fragmentation, it results in a proportionally increased computational cost, making LLM-based applications more resource-intensive and expensive. Moreover, this inefficiency heightens the environmental toll by requiring more energy to process longer token sequences.

Addressing these inefficiencies requires innovations in tokenization algorithms that can better accommodate diverse linguistic structures. By reducing token fragmentation, we can lower the FLOP count and, consequently, the carbon emissions associated with LLMs. This would contribute to global efforts to mitigate the climate impact of AI technologies and promote linguistic equity by making language models more accessible and cost-effective for speakers of diverse languages.

While the environmental impact of LLMs is influenced by various factors---such as hardware energy efficiency and data center infrastructure---the efficiency of the models themselves and their related components remains a key determinant. As LLMs continue to scale in size and usage, addressing tokenization inefficiencies becomes increasingly urgent. Improvements in this area would benefit everyone by reducing the global carbon emissions associated with LLMs. Therefore, one of the pathways toward creating more sustainable and equitable LLMs is through optimizing both the technical and linguistic aspects of tokenization. Continuous efforts to refine tokenization algorithms, aimed at reducing the total number of tokens these tokenizers produce across all languages, including English, can significantly mitigate the climate impact of AI technologies.

\subsection{Delocalization of Cost to Access}

Besides the LLM-intrinsic issues detailed in this paper, there are additional factors that exacerbate access challenges and inequalities in cloud-based AI services. A major concern is the delocalization of costs, where global service providers, relying on major currencies, create substantial barriers for users in low- and lower-middle-income countries.

Providers such as OpenAI, AWS, Microsoft Azure, and Google Cloud Platform (GCP) typically charge in major currencies, primarily the US dollar (USD). While this simplifies global operations, it poses significant challenges for users whose local currencies fluctuate against the USD. Forex volatility can sharply increase cloud service costs for businesses and individuals in economically unstable regions.

This issue is particularly acute in low- and lower-middle-income countries, where currency instability is often higher. When local currencies depreciate against the USD, the costs of accessing LLM services through cloud platforms can rise dramatically. For instance, a company in such a country may find that weakening currency renders once-affordable services prohibitively expensive, further straining already-limited resources.

These rising costs widen the economic gap in accessing advanced technologies. Businesses and individuals in low- and lower-middle-income countries are already disadvantaged by higher tokenization premiums for less commonly spoken languages. The added impact and uncertainty of forex-driven cost hikes exacerbate this divide, presenting dual challenges: elevated premiums and unpredictable service costs.

Such financial pressures limit access and potential of businesses in these regions from innovating or competing globally. While wealthier nations enjoy stable, affordable cloud services, lower-income regions face cost inconsistencies that hinder their ability to engage with cutting-edge AI technologies.

The long-term effects of rising cloud service costs are substantial. Many businesses in low- and lower-middle-income countries may struggle to afford these services, hampering their ability to leverage LLMs and broader AI tools. This financial strain could discourage startups, schools, and local businesses from adopting AI, stifling innovation and fostering a ``brain drain" as local talent turns away from costly AI projects, further hindering technological growth in these regions.

Additionally, businesses may be driven to seek out cheaper, lower-quality options, which could offer reduced performance and scalability. This would further impede these regions from fully participating in the global digital economy, reinforcing economic disparities.

To address the combined challenges of forex fluctuations and tokenization premiums, service providers and policymakers should explore solutions to make cloud services more affordable in low- and lower-middle-income countries. One potential strategy is pegging costs to local currencies, which could stabilize pricing and ensure more consistent service fees.

Governments could also play a role by offering subsidies or incentives to businesses that rely heavily on cloud services. Encouraging the growth of regional cloud providers that bill in local currencies and are less vulnerable to international currency fluctuations could provide a more stable and cost-effective solution for users in low- and lower-middle-income economies.

\subsection{A \textit{Subtle} Risk of Inaction}

A substantial yet frequently unacknowledged risk of not improving LLM performance for low-resource languages is the potential contamination of the internet with subpar or incorrect content in these languages \cite{shumailov_ai_2024}. As LLMs continue generating text that appears to come from low-resource languages, but are trained on limited and unreliable data, the chances of producing flawed translations or erroneous outputs increase. This risk grows when models cannot accurately capture the nuances and complexities of these languages, leading to outputs that may seem correct but are often inaccurate or misleading. When these inaccuracies are disseminated online, they contribute to an expanding pool of low-quality linguistic data, which can further exacerbate the performance challenges of LLMs in future updates.

This concern is particularly pressing because LLMs rely heavily on internet-sourced data for training. If inaccurate or subpar content in low-resource languages proliferates online, it risks being incorporated into future training datasets, thereby perpetuating a cycle of substandard performance. As more flawed material is produced and shared, LLMs will find it increasingly difficult to generate accurate text in these languages, resulting in a self-perpetuating feedback loop.

Moreover, if speakers of these languages depend on AI-generated content for education, communication, or business, the consequences could be significant. Errors or mistranslations may lead to misunderstandings, spread misinformation, and erode trust in AI among low-resource language users, further widening the digital divide and limiting their participation in the global digital ecosystem.

Failing to improve LLMs for low-resource languages risks degrading data quality and perpetuating poor performance, further marginalizing these languages. Enhancing AI-generated content and ensuring that LLMs are trained on accurate and culturally relevant data is essential.

\section{Conclusion}\label{conclusion}

The increasing capabilities and utility of Large Language Models (LLMs) present a unique opportunity to bridge global knowledge and access gaps. However, for speakers of low-resource languages, this potential remains largely unrealized, resulting in a double jeopardy: they face both elevated tokenization costs and inferior model performance. Although advancements like GPT-4o have minimized tokenization premiums to some extent, these gains are unevenly distributed, continuing to exclude speakers from lower-income and lower-middle-income nations. About 1.5 billion speakers in lower-middle-income countries face tokenization costs 4 to 6 times higher than those for English. This dual burden of cost and inefficiency exacerbates existing inequities, limiting access to advanced AI technologies for already marginalized groups.

In addition to these intrinsic challenges, external economic factors, such as currency fluctuations, further increase the cost of cloud-based AI services for businesses in economically unstable areas. This cost shift widens the global economic gap, limiting access to cutting-edge technologies and stifling innovation in lower-income economies, thus deepening the digital divide.

Moreover, the environmental cost of inefficient tokenization, especially for low-resource languages, compounds the challenge. As LLMs scale, their energy consumption and carbon emissions grow, particularly when models process unnecessarily long token sequences due to inefficiencies. Reducing tokenization inefficiencies across all languages, including well-resourced ones like English, is critical to minimizing the climate impact of AI technologies. While long-context LLMs are gaining traction, there is an urgent need for more efficient tokenization algorithms that can process the same amount of information using shorter contexts. This shift could reduce computational demands, but ensuring LLM performance is maintained or enhanced will be crucial in this new tokenization paradigm. By addressing these inefficiencies, we can directly reduce the environmental footprint of AI, ensuring its development remains sustainable.

If LLMs for low-resource languages are not improved, we risk creating a feedback loop where poor model performance leads to degraded training data, further marginalizing these languages. Proactive steps are essential to ensure that speakers of low-resource languages can fully access and benefit from AI technologies. Both technological advancements and policy initiatives must be pursued to provide equitable access to AI systems. This requires addressing the economic and linguistic disparities and the environmental costs of large-scale AI deployment, ensuring that no group is left behind in the digital revolution while safeguarding the planet’s resources.

Considering the multidimensional challenges surrounding LLMs, addressing these issues requires immediate and concerted efforts from all stakeholders. Companies, organizations, and academic institutions must intensify their investments in research and development to optimize tokenization strategies that effectively address fragmentation issues, or explore alternative methods for interfacing raw data with LLMs. Governments, in turn, need to address existing gaps that could hinder the localized application of these technologies, particularly in regions where economic and infrastructural disparities limit access to affordable cloud services. Furthermore, governments should incorporate provisions into their AI roadmaps to ensure that resources, including those in local languages, are made AI-ready, facilitating their use in training and integration with LLMs—\textbf{Data for AI}. This includes supporting initiatives to digitize diverse materials---such as recordings, books, and other cultural resources---and investing in data curation to create high-quality, semantically rich datasets that foster the development of inclusive AI models and applications. Multilateral development organizations must support governments in these efforts to ensure success.

\bibliography{sn-article} 

\backmatter

\bmhead{Supplementary information}

The code for this paper is accessible on GitHub: \url{https://github.com/worldbank/double-jeopardy-in-llms}.

\bmhead{Disclaimer and Disclosure of AI Use}

The findings, interpretations, and conclusions expressed in this paper are entirely those of the authors. They do not necessarily represent the views of the International Bank for Reconstruction and Development/World Bank and its affiliated organizations, or those of the Executive Directors of the World Bank or the governments they represent.

This work used AI tools at various stages, including assessing the performance of LLMs on multiple language translation tasks using OpenAI's GPT-4 Turbo and GPT-4o models (APIs). In addition, Microsoft Co-Pilot and ChatGPT were employed to enhance the manuscript's readability.










\begin{appendices}

    \section{Additional plots}\label{secA1}

    \begin{figure}[t]
    \begin{center}
        \centerline{\includegraphics[width=\columnwidth]{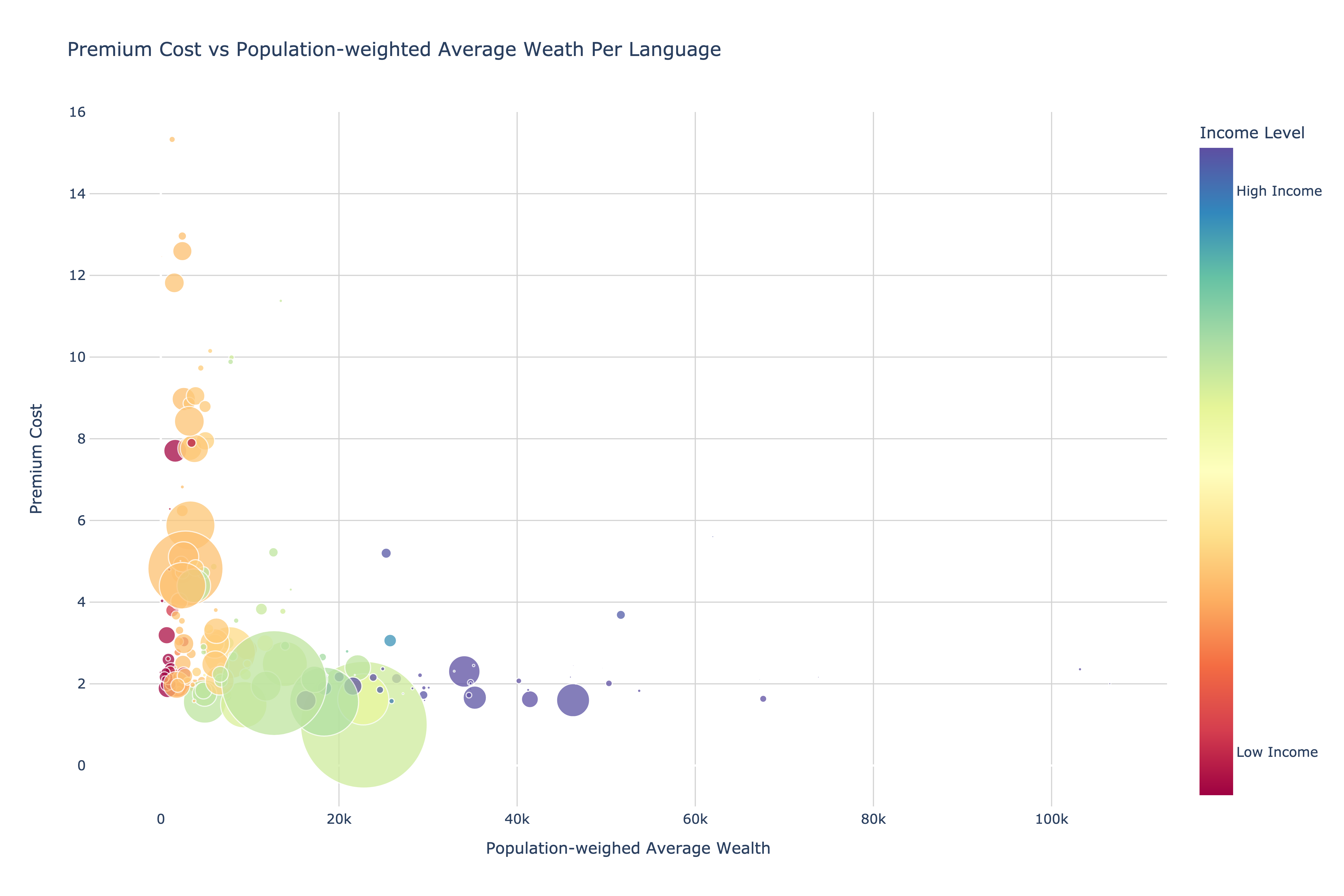}}
        \caption{Full extent of population weighted average wealth.}
        \label{fig:lang-premium-cost}
    \end{center}
\end{figure}


\begin{figure}[t]
    \begin{center}
        \centerline{\includegraphics[width=\columnwidth]{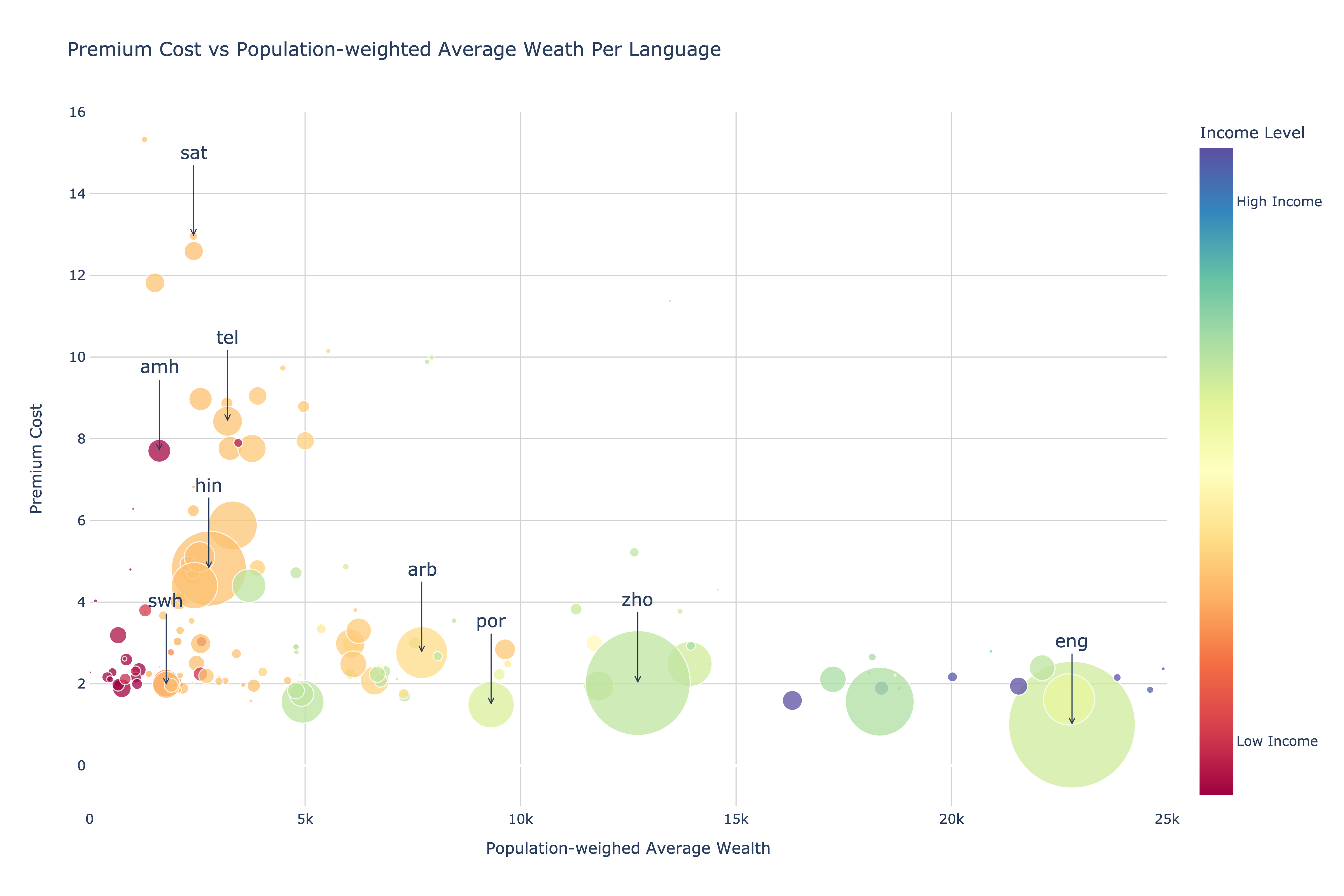}}
        \caption{Population weighted average wealth cut off at \$25k.}
        \label{fig:lang-premium-cost-25K}
    \end{center}
\end{figure}




\end{appendices}



\end{document}